\documentclass[10pt,twocolumn,letterpaper]{article}

\usepackage{conf}
\usepackage{times}
\usepackage{epsfig}
\usepackage{graphicx}
\usepackage{amsmath}
\usepackage{amssymb}

\usepackage{enumitem}
\usepackage{subcaption}
\usepackage[table]{xcolor}
\usepackage{multirow}
\usepackage{hhline}
\usepackage{array}
\usepackage{float}
\usepackage{subcaption}
\usepackage{subfiles}
\usepackage[pagebackref=true,breaklinks=true,letterpaper=true,colorlinks,bookmarks=false]{hyperref}
\graphicspath{{figures/}}

\conffinalcopy %

\def\confPaperID{} %

\begin{document}

\title{EdgeConnect: Generative Image Inpainting with Adversarial Edge Learning}

\author{
	Kamyar Nazeri, \,\,
	Eric Ng, \,\,
	Tony Joseph, \,\,
	Faisal Z. Qureshi, \,\,
	Mehran Ebrahimi \vspace{3pt}\\
	Faculty of Science, University of Ontario Institute of Technology, Canada\\
	{\tt\small \{kamyar.nazeri, eric.ng, tony.joseph, faisal.qureshi, mehran.ebrahimi\}@uoit.ca}
}

\maketitle


\begin{abstract}

	Over the last few years, deep learning techniques have yielded significant improvements in image inpainting. However, many of these techniques fail to reconstruct reasonable structures as they are commonly over-smoothed and/or blurry. This paper develops a new approach for image inpainting that does a better job of reproducing filled regions exhibiting fine details. We propose a two-stage adversarial model EdgeConnect that comprises of an edge generator followed by an image completion network. The edge generator hallucinates edges of the missing region (both regular and irregular) of the image, and the image completion network fills in the missing regions using hallucinated edges as \textit{a priori}. We evaluate our model end-to-end over the publicly available datasets CelebA, Places2, and Paris StreetView, and show that it outperforms current state-of-the-art techniques quantitatively and qualitatively.

\end{abstract}
\section{Introduction}

	\begin{figure}[t]
		\includegraphics[width=0.45\textwidth]{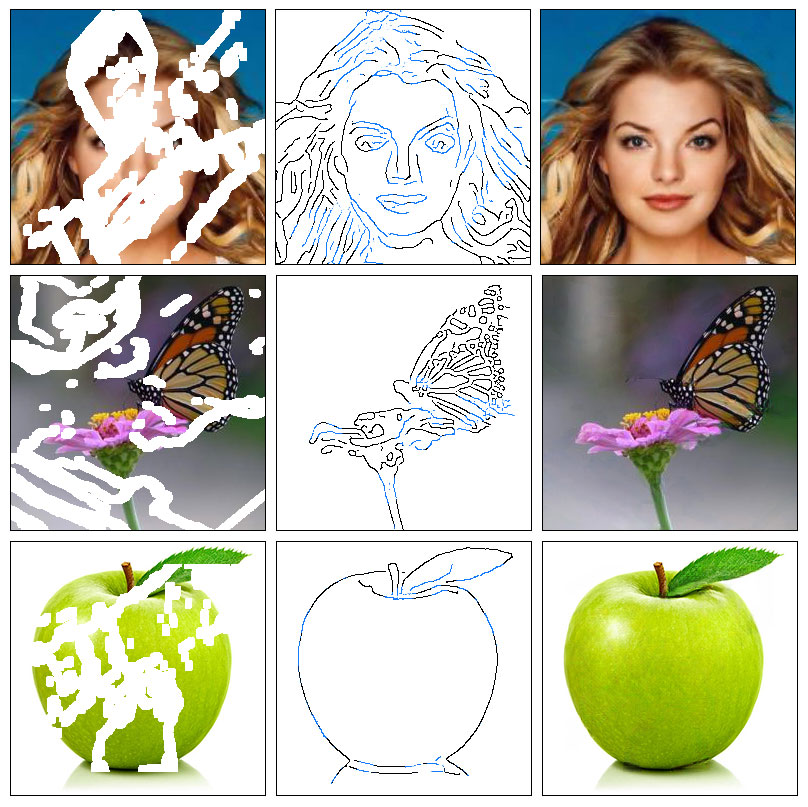}
		\caption{(Left) Input images with missing regions. The missing regions are depicted in white. (Center) Computed edge masks. Edges drawn in black are computed (for the available regions) using Canny edge detector; whereas edges shown in blue are hallucinated (for the missing regions) by the edge generator network. (Right) Image inpainting results of the proposed approach.}
		\label{fig:intro}
	\end{figure}
	Image inpainting, or image completion, involves filling in missing regions of an image. It is an important step in many image editing tasks. It can, for example, be used to fill in the holes left after removing unwanted objects from an image. Humans have an uncanny ability to zero in on visual inconsistencies. Consequently, the filled regions must be perceptually plausible. Among other things, the lack of fine structure in the filled region is a giveaway that something is amiss, especially when the rest of the image contain sharp details. The work presented in this paper is motivated by our observation that many existing image inpainting techniques generate over-smoothed and/or blurry regions, failing to reproduce fine details.
	
	We divide image inpainting into a two-stage process (Figure \ref{fig:intro}): edge generation and image completion. Edge generation is solely focused on hallucinating edges in the missing regions. The image completion network uses the hallucinated edges and estimates RGB pixel intensities of the missing regions. Both stages follow an adversarial framework \cite{goodfellow2014generative} to ensure that the hallucinated edges and the RGB pixel intensities are visually consistent. Both networks incorporate losses based on deep features to enforce perceptually realistic results.
	
	Like most computer vision problems, image inpainting predates the wide-spread use of deep learning techniques. Broadly speaking, traditional approaches for image inpainting can be divided into two groups: diffusion-based and patch-based. Diffusion-based methods propagate background data into the missing region by following a diffusive process typically modeled using differential operators \cite{bertalmio2000image, esedoglu2002digital, liu2007image, ballester2001filling}. Patch-based methods, on the other hand, fill in missing regions with patches from a collection of source images that maximize patch similarity \cite{darabi2012image,huang2014image}. These methods, however, do a poor job of reconstructing complex details that may be local to the missing region.
	
	More recently deep learning approaches have found remarkable success at the task of image inpainting. These schemes fill the missing pixels using learned data distribution. They are able to generate coherent structures in the missing regions, a feat that was nearly impossible for traditional techniques. While these approaches are able to generate missing regions with meaningful structures, the generated regions are often blurry or suffer from artifacts, suggesting that these methods struggle to reconstruct high frequency information accurately.
	
	Then, how does one force an image inpainting network to generate fine details? Since image structure is well-represented in its edge mask, we show that it is possible to generate superior results by conditioning an image inpainting network on edges in the missing regions. Clearly, we do not have access to edges in the missing regions. Rather, we train an edge generator that hallucinates edges in these areas. Our approach of ``lines first, color next'' is partly inspired by our understanding of how artists work \cite{eitz2012humans}. ``\emph{In line drawing, the lines not only delineate and define spaces and shapes; they also play a vital role in the composition}'', says Betty Edwards, highlights the importance of sketches from an artistic viewpoint \cite{edwards2012drawing}. Edge recovery, we suppose, is an easier task than image completion. Our proposed model essentially decouples the recovery of high and low-frequency information of the inpainted region.
	
	We evaluate our proposed model on standard datasets CelebA \cite{liu2015faceattributes}, Places2 \cite{zhou2017places}, and Paris StreetView \cite{doersch2012makes}. We compare the performance of our model against current state-of-the-art schemes. Furthermore, we provide results of experiments carried out to study the effects of edge information on the image inpainting task. Our paper makes the following contributions:
	\begin{itemize}[noitemsep,topsep=0pt]
		\item An edge generator capable of hallucinating edges in missing regions given edges and grayscale pixel intensities of the rest of the image.
		\item An image completion network that combines edges in the missing regions with color and texture information of the rest of the image to fill the missing regions.
		\item An end-to-end trainable network that combines edge generation and image completion to fill in missing regions exhibiting fine details.
	\end{itemize}
	We show that our model can be used in some common image editing applications, such as object removal and scene generation. Our source code is available at:\\ \small\url{https://github.com/knazeri/edge-connect}

\section{Related Work}

	\textit{Diffusion-based} methods propagate neighboring information into the missing regions \cite{bertalmio2000image, ballester2001filling}. \cite{esedoglu2002digital} adapted the Mumford-Shah segmentation model for image inpainting by introducing Euler's Elastica. However, reconstruction is restricted to locally available information for these diffusion-based methods, and these methods fail to recover meaningful structures in the missing regions. These methods also cannot adequately deal with large missing regions.
	
	\textit{Patch-based} methods fill in missing regions (\ie, targets) by copying information from similar regions (\ie, sources) of the same image (or a collection of images). Source regions are often blended into the target regions to minimize discontinuities \cite{darabi2012image, huang2014image}. These methods are computationally expensive since similarity scores must be computed for every target-source pair. PatchMatch \cite{barnes2009patchmatch} addressed this issue by using a fast nearest neighbor field algorithm. These methods, however, assume that the texture of the inpainted region can be found elsewhere in the image. This assumption does not always hold. Consequently, these methods excel at recovering highly patterned regions such as background completion but struggle at reconstructing patterns that are locally unique.
	
	One of the first \textit{deep learning} methods designed for image inpainting is context encoder \cite{pathak2016context}, which uses an encoder-decoder architecture. The encoder maps an image with missing regions to a low-dimensional feature space, which the decoder uses to construct the output image. However, the recovered regions of the output image often contain visual artifacts and exhibit blurriness due to the information bottleneck in the channel-wise fully connected layer. This was addressed by Iizuka \etal \cite{iizuka2017globally} by reducing the number of downsampling layers, and replacing the channel-wise fully connected layer with a series of dilated convolution layers \cite{yu2016multi}. The reduction of downsampling layers are compensated by using varying dilation factors. However, training time was increased significantly\footnote{Model by \cite{iizuka2017globally} required two months of training over four GPUs.} due to extremely sparse filters created using large dilation factors. Yang \etal \cite{yang2017high} uses a pre-trained VGG network \cite{simonyan2014very} to improve the output of the context-encoder, by minimizing the feature difference of image background. This approach requires solving a multi-scale optimization problem iteratively, which noticeably increases computational cost during inference time. Liu \etal \cite{Liu_2018_ECCV} introduced ``partial convolution'' for image inpainting, where convolution weights are normalized by the mask area of the window that the convolution filter currently resides over. This effectively prevents the convolution filters from capturing too many zeros when they traverse over the incomplete region.
	
	Recently, several methods were introduced by providing additional information prior to inpainting. Yeh \etal \cite{yeh2017semantic} trains a GAN for image inpainting with uncorrupted data. During inference, back-propagation is employed for $1,500$ iterations  to find the representation of the corrupted image on a uniform noise distribution. However, the model is slow during inference since back-propagation must be performed for every image it attempts to recover. Dolhansky and Ferrer \cite{dolhansky2018eye} demonstrate the importance of exemplar information for inpainting. Their method is able to achieve both sharp and realistic inpainting results. Their method, however, is geared towards filling in missing eye regions in frontal human face images. It is highly specialized and does not generalize well. Contextual Attention \cite{yu2018generative} takes a two-step approach to the problem of image inpainting. First, it produces a coarse estimate of the missing region. Next, a refinement network sharpens the result using an attention mechanism by searching for a collection of background patches with the highest similarity to the coarse estimate. \cite{song2018contextual} takes a similar approach and introduces a ``patch-swap'' layer which replaces each patch inside the missing region with the most similar patch on the boundary. These schemes suffer from two limitations: 1) the refinement network assumes that the coarse estimate is reasonably accurate, and 2) these methods cannot handle missing regions with arbitrary shapes. Free-form inpainting method proposed in \cite{yu2018free} is perhaps closest in spirit to our scheme. It uses hand-drawn sketches to guide the inpainting process. Our method does away with hand-drawn sketches and instead learns to hallucinate edges in the missing regions.
	
	\subsection{Image-to-Edges vs. Edges-to-Image}
	The inpainting technique proposed in this paper subsumes two disparate computer vision problems: Image-to-Edges and Edges-to-Image. There is a large body of literature that addresses ``Image-to-Edges'' problems \cite{bertasius2015deepedge, dollar2006supervised, li2016unsupervised, liu2017richer}. Canny edge detector, an early scheme for constructing edge maps, for example, is roughly 30 years old \cite{canny1986computational}. Doll{\'a}r and Zitnikc \cite{dollar2015fast} use \emph{structured learning} \cite{nowozin2011structured} on random decision forests to predict local edge masks. Holistically-nested Edge Detection (HED) \cite{xie2015holistically} is a fully convolutional network that learns edge information based on its importance as a feature of the overall image. In our work, we train on edge maps computed using Canny edge detector. We explain this in detail in Section \ref{edge} and Section \ref{ablation}.
	
	Traditional ``Edges-to-Image'' methods typically follow a bag-of-words approach, where image content is constructed through a pre-defined set of keywords. These methods, however, are unable to accurately construct fine-grained details especially near object boundaries. Scribbler \cite{sangkloy2017scribbler} is a learning-based model where images are generated using line sketches as the input. The results of their work possess an art-like quality, where color distribution of the generated result is guided by the use of color in the input sketch. Isola \etal \cite{isola2017image} proposed a conditional GAN framework \cite{mirza2014conditional}, called pix2pix, for image-to-image translation problems. This scheme can use available edge information as \textit{a priori}. CycleGAN \cite{zhu2017unpaired} extends this framework and finds a reverse mapping back to the original data distribution. This approach yields superior results since the aim is to learn the inverse of the forward mapping.

\section{EdgeConnect}

	\begin{figure*}
		\centering
		\includegraphics[height=.16\textheight]{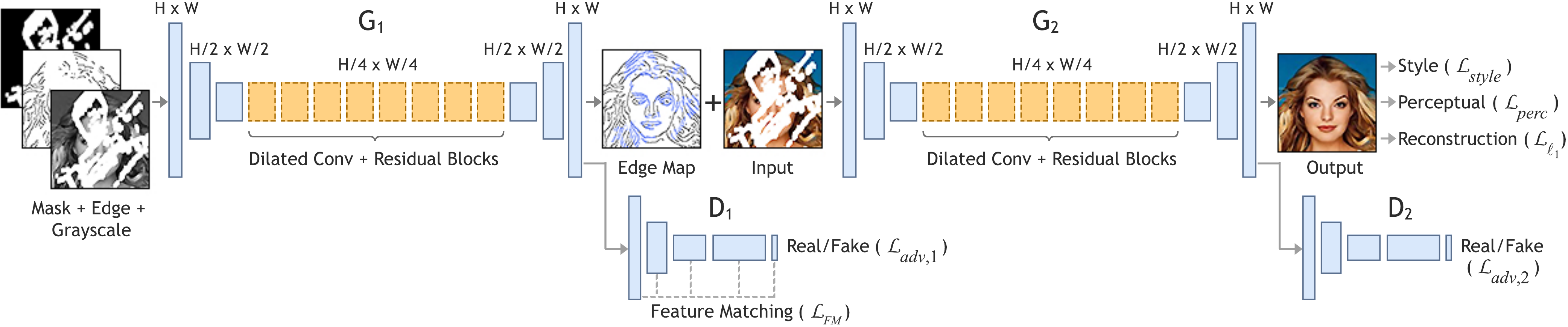}
	\caption{Summary of our proposed method. Incomplete grayscale image and edge map, and mask are the inputs of $G_1$ to predict the full edge map. Predicted edge map and incomplete color image are passed to $G_2$ to perform the inpainting task.}
		\label{fig:arch}
	\end{figure*}
	We propose an image inpainting network that consists of two stages: 1) edge generator, and 2) image completion network (Figure \ref{fig:arch}). Both stages follow an adversarial model \cite{goodfellow2014generative}, \ie each stage consists of a generator/discriminator pair. Let $G_1$ and $D_1$ be the generator and discriminator for the edge generator, and $G_2$ and $D_2$ be the generator and discriminator for the image completion network, respectively. To simplify notation, we will use these symbols also to represent the function mappings of their respective networks.

	Our generators follow an architecture similar to the method proposed by Johnson \etal \cite{johnson2016perceptual}, which has achieved impressive results for style transfer, super-resolution \cite{sajjadi2017enhancenet, gondal2018unreasonable}, and image-to-image translation \cite{zhu2017unpaired}. Specifically, the generators consist of encoders that down-sample twice, followed by eight residual blocks \cite{he2016deep} and decoders that up-sample images back to the original size. Dilated convolutions with a dilation factor of two are used instead of regular convolutions in the residual layers, resulting in a receptive field of $205$ at the final residual block. For discriminators, we use a $70 \times 70$ PatchGAN \cite{isola2017image, zhu2017unpaired} architecture, which determines whether or not overlapping image patches of size $70 \times 70$ are real. We use instance normalization \cite{ulyanov2017improved} across all layers of the network\footnote{The details of our architecture are in appendix \ref{app:netwrok}}.

	\subsection{Edge Generator}
	Let $\mathbf{I}_{gt}$ be ground truth images. Their edge map and grayscale counterpart will be denoted by $\mathbf{C}_{gt}$ and $\mathbf{I}_{gray}$, respectively. In the edge generator, we use the masked grayscale image $\tilde{\mathbf{I}}_{gray} = \mathbf{I}_{gray} \odot (\mathbf{1} - \mathbf{M})$ as the input, its edge map $\tilde{\mathbf{C}}_{gt} = \mathbf{C}_{gt} \odot (\mathbf{1} - \mathbf{M})$, and image mask $\mathbf{M}$ as a pre-condition (1 for the missing region, 0 for background). Here, $\odot$ denotes the Hadamard product. The generator predicts the edge map for the masked region
	\begin{equation}
		\mathbf{C}_{pred} = G_1 \left( \tilde{\mathbf{I}}_{gray}, \tilde{\mathbf{C}}_{gt}, \mathbf{M} \right).
	\end{equation}
	We use $\mathbf{C}_{gt}$ and $\mathbf{C}_{pred}$ conditioned on $\mathbf{I}_{gray}$ as inputs of the discriminator that predicts whether or not an edge map is real. The network is trained with an objective comprised of an adversarial loss and feature-matching loss \cite{wang2018high}
	\begin{equation}
		\min_{G_1} \max_{D_1} \mathcal{L}_{G_1} = \min_{G_1} \left( \lambda_{adv,1} \max_{D_1} \left( \mathcal{L}_{adv,1} \right) + \lambda_{FM} \mathcal{L}_{FM} \right)
	\end{equation}
	where $\lambda_{adv,1}$ and $\lambda_{FM}$ are regularization parameters. The adversarial loss is defined as
	\begin{multline}
		\mathcal{L}_{adv,1} =\mathbb{E}_{(\mathbf{C}_{gt},\mathbf{I}_{gray})} \left[ \log D_1(\mathbf{C}_{gt},\mathbf{I}_{gray}) \right] \\
		+ \mathbb{E}_{\mathbf{I}_{gray}} \log \left[ 1 - D_1(\mathbf{C}_{pred}, \mathbf{I}_{gray}) \right].
		\label{eq:g1_adv}
	\end{multline}
	The feature-matching loss $\mathcal{L}_{FM}$ compares the activation maps in the intermediate layers of the discriminator. This stabilizes the training process by forcing the generator to produce results with representations that are similar to real images. This is similar to perceptual loss \cite{johnson2016perceptual,gatys2016image,gatys2015texture}, where activation maps are compared with those from the pre-trained VGG network. However, since the VGG network is not trained to produce edge information, it fails to capture the result that we seek in the initial stage. The feature matching loss $\mathcal{L}_{FM}$ is defined as
	\begin{equation}
		\mathcal{L}_{FM} = \mathbb{E} \left[ \sum_{i=1}^L \frac{1}{N_i} \left\lVert D^{(i)}_1 (\mathbf{C}_{gt}) - D^{(i)}_1 \left( \mathbf{C}_{pred} \right) \right\rVert_1 \right],
	\end{equation}
	where $L$ is the final convolution layer of the discriminator, $N_i$ is the number of elements in the $i$'th activation layer, and $D^{(i)}_1$ is the activation in the $i$'th layer of the discriminator. Spectral normalization (SN) \cite{miyato2018spectral} further stabilizes training by scaling down weight matrices by their respective largest singular values, effectively restricting the Lipschitz constant of the network to one. Although this was originally proposed to be used only on the discriminator, recent works \cite{zhang2018self, odena2018generator} suggest that generator can also benefit from SN by suppressing sudden changes of parameter and gradient values. Therefore, we apply SN to both generator and discriminator. Spectral normalization was chosen over Wasserstein GAN (WGAN), \cite{arjovsky2017wasserstein} as we found that WGAN was several times slower in our early tests. Note that only $\mathcal{L}_{adv,1}$ is maximized over $D_1$ since $D_1$ is used to retrieve activation maps for $\mathcal{L}_{FM}$. For our experiments, we choose $\lambda_{adv,1} = 1$ and $\lambda_{FM} = 10$.
	\begin{figure*}[t]
		\centering
		\begin{subfigure}[c]{.14\textwidth}
			\centering
			\includegraphics[width=\textwidth]{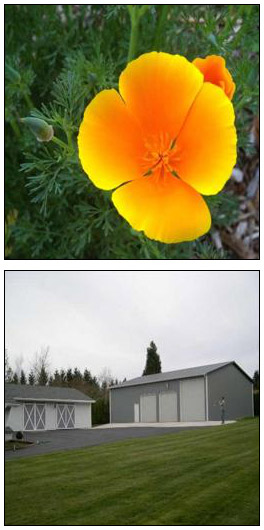}
			\caption{\footnotesize{Ground Truth}}
		\end{subfigure}
		\begin{subfigure}[c]{.14\textwidth}
			\centering
			\includegraphics[width=\textwidth]{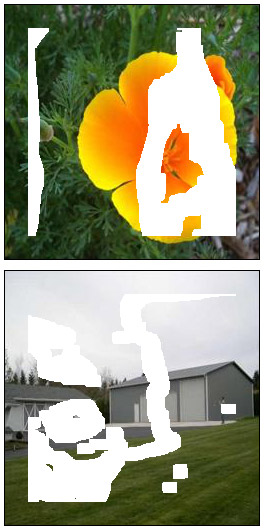}
			\caption{\footnotesize{Masked Image}}
		\end{subfigure}
		\begin{subfigure}[c]{.14\textwidth}
			\centering
			\includegraphics[width=\textwidth]{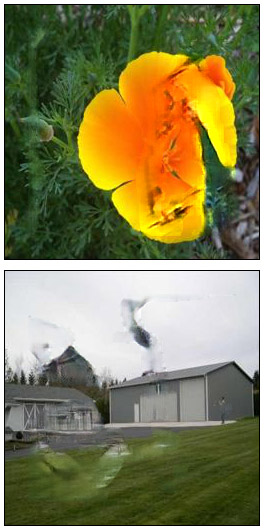}
			\caption{\footnotesize{Yu \etal \cite{yu2018generative}}}
		\end{subfigure}
		\begin{subfigure}[c]{.14\textwidth}
			\centering
			\includegraphics[width=\textwidth]{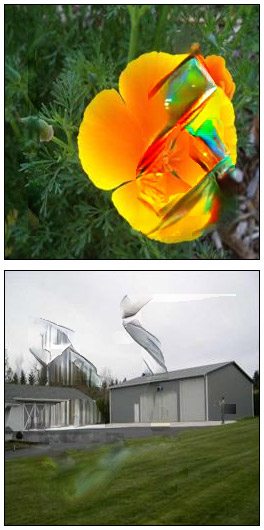}
			\caption{\footnotesize{Iizuka \etal \cite{iizuka2017globally}}}
		\end{subfigure}
		\begin{subfigure}[c]{.14\textwidth}
			\centering
			\includegraphics[width=\textwidth]{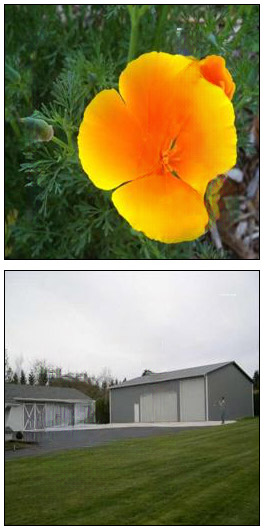}
			\caption{\footnotesize{Ours}}
		\end{subfigure}
		\begin{subfigure}[c]{.14\textwidth}
			\centering
			\includegraphics[width=\textwidth]{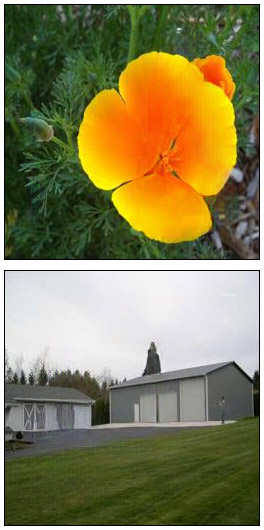}
			\caption{\footnotesize{Ours (Canny)}}
		\end{subfigure} 
		\caption{Comparison of qualitative results with existing models. (a) Ground Truth Image. (b) Ground Truth with Mask. (c) Yu \etal \cite{yu2018generative}. (d) Iizuka \etal \cite{iizuka2017globally}. (e) Ours (end-to-end). (f) Ours ($G_2$ only with Canny $\sigma = 2$).}
		\label{fig:compare}
	\end{figure*}
	\subsection{Image Completion Network}
	The image completion network uses the incomplete color image $\tilde{\mathbf{I}}_{gt} = \mathbf{I}_{gt} \odot (\mathbf{1-M})$ as input, conditioned using a composite edge map $\mathbf{C}_{comp}$. The composite edge map is constructed by combining the background region of ground truth edges with generated edges in the corrupted region from the previous stage, \ie $\mathbf{C}_{comp} = \mathbf{C}_{gt} \odot (\mathbf{1-M}) + \mathbf{C}_{pred} \odot \mathbf{M}$. The network returns a color image $\mathbf{I}_{pred}$, with missing regions filled in, that has the same resolution as the input image:
	\begin{equation}
		\mathbf{I}_{pred} = G_2 \left( \tilde{\mathbf{I}}_{gt}, \mathbf{C}_{comp} \right).
	\end{equation}
	This is trained over a joint loss that consists of an $\ell_1$ loss, adversarial loss, perceptual loss, and style loss. To ensure proper scaling, the $\ell_1$ loss is normalized by the mask size. The adversarial loss is defined similar to Eq. \ref{eq:g1_adv}, as
	\begin{multline}
		\mathcal{L}_{adv,2} =\mathbb{E}_{(\mathbf{I}_{gt},\mathbf{C}_{comp})} \left[ \log D_2(\mathbf{I}_{gt},\mathbf{C}_{comp}) \right] \\
		+ \mathbb{E}_{\mathbf{C}_{comp}} \log \left[ 1 - D_2(\mathbf{I}_{pred}, \mathbf{C}_{comp}) \right].
	\end{multline}
	We include the two losses proposed in \cite{gatys2016image, johnson2016perceptual} commonly known as perceptual loss $\mathcal{L}_{perc}$ and style loss $\mathcal{L}_{style}$. As the name suggests, $\mathcal{L}_{perc}$ penalizes results that are not perceptually similar to labels by defining a distance measure between activation maps of a pre-trained network. Perceptual loss is defined as
	\begin{equation}
		\mathcal{L}_{perc} = \mathbb{E} \left[ \sum_i \frac{1}{N_i} \left\lVert \phi_i (\mathbf{I}_{gt}) - \phi_i (\mathbf{I}_{pred}) \right \rVert_1 \right]
	\end{equation}
	where $\phi_i$ is the activation map of the $i$'th layer of a pre-trained network. For our work, $\phi_i$ corresponds to activation maps from layers $\tt{relu1\_1, ~relu2\_1, ~relu3\_1, ~relu4\_1}$ and $\tt{relu5\_1}$ of the VGG-19 network pre-trained on the ImageNet dataset \cite{russakovsky2015imagenet}. These activation maps are also used to compute style loss which measures the differences between covariances of the activation maps. Given feature maps of sizes $C_j \times H_j \times W_j$, style loss is computed by 
	\begin{equation}
		\mathcal{L}_{style} = \mathbb{E}_j \left[ \lVert G_j^{\phi} (\tilde{\mathbf{I}}_{pred}) - G_j^{\phi} (\tilde{\mathbf{I}}_{gt}) \rVert_1 \right]
	\end{equation}
	where $G_j^{\phi}$ is a $C_j \times C_j$ Gram matrix constructed from activation maps $\phi_j$. We choose to use style loss as it was shown by Sajjadi \etal \cite{sajjadi2017enhancenet} to be an effective tool to combat ``checkerboard'' artifacts caused by transpose convolution layers \cite{odena2016deconvolution}. Our overall loss function is
	\begin{equation}
		\mathcal{L}_{G_2} = \lambda_{\ell_1} \mathcal{L}_{\ell_1} + \lambda_{adv,2} \mathcal{L}_{adv,2} + \lambda_p \mathcal{L}_{perc} + \lambda_s \mathcal{L}_{style}.
	\end{equation}
	For our experiments, we choose $\lambda_{\ell_1} = 1$, $\lambda_{adv,2} = \lambda_p = 0.1$, and $\lambda_s = 250$. We noticed that the training time increases significantly if spectral normalization is included. We believe this is due to the network becoming too restrictive with the increased number of terms in the loss function. Therefore we choose to exclude spectral normalization from the image completion network.

\section{Experiments}

	\subsection{Edge Information and Image Masks} \label{edge}
	To train $G_1$, we generate training labels (\ie edge maps) using Canny edge detector. The sensitivity of Canny edge detector is controlled by the standard deviation of the Gaussian smoothing filter $\sigma$. For our tests, we empirically found that $\sigma \approx 2$ yields the best results (Figure \ref{fig:sig}). In Section \ref{ablation}, we investigate the effect of the quality of edge maps on the overall image completion.
	
	For our experiments, we use two types of image masks: regular and irregular. Regular masks are square masks of fixed size (25\% of total image pixels) centered at a random location within the image. We obtain irregular masks from the work of Liu \etal \cite{Liu_2018_ECCV}. Irregular masks are augmented by introducing four rotations ($0^{\circ}, 90^{\circ}, 180^{\circ}, 270^{\circ}$) and a horizontal reflection for each mask. They are classified based on their sizes relative to the entire image in increments of $10\%$ (\eg, 0-10\%, 10-20\%, \etc).
	\subsection{Training Setup and Strategy}
	Our proposed model is implemented in PyTorch. The network is trained using $256 \times 256$ images with a batch size of eight. The model is optimized using Adam optimizer \cite{kingma2014adam} with $\beta_1 = 0$ and $\beta_2 = 0.9$. Generators $G_1, G_2$ are trained separately using Canny edges with learning rate $10^{-4}$ until the losses plateau. We lower the learning rate to $10^{-5}$ and continue to train $G_1$ and $G_2$ until convergence. Finally, we fine-tune the networks by removing $D_1$, then train $G_1$ and $G_2$ end-to-end with learning rate $10^{-6}$ until convergence. Discriminators are trained with a learning rate one tenth of the generators'.

\section{Results}

	Our proposed model is evaluated on the datasets CelebA \cite{liu2015faceattributes}, Places2 \cite{zhou2017places}, and Paris StreetView \cite{doersch2012makes}. Results are compared against the current state-of-the-art methods both qualitatively and quantitatively.
	\subsection{Qualitative Comparison}
\begin{figure}[ht!]
		\centering
		\includegraphics[width=.46\textwidth]{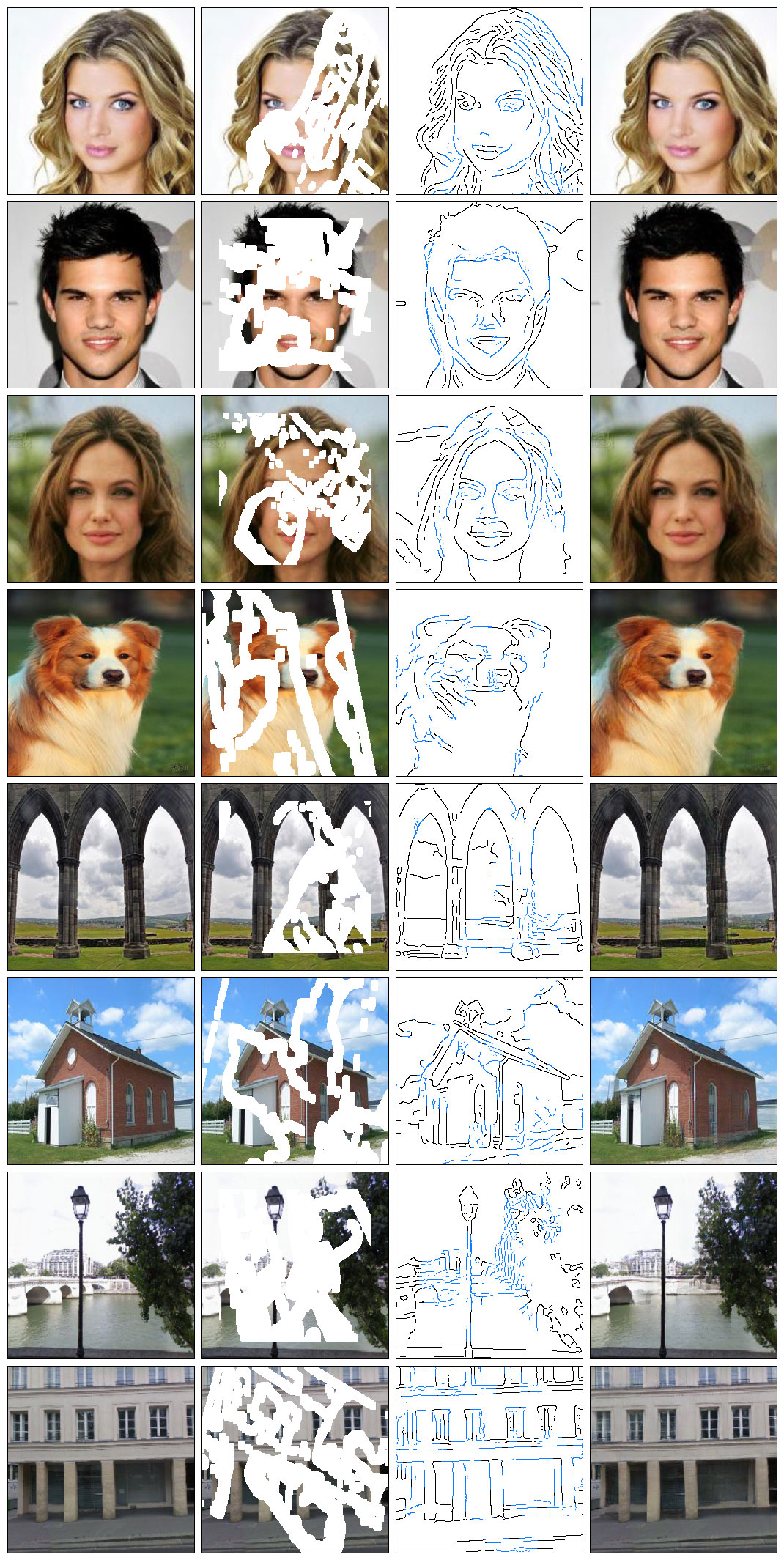}
		\caption{(Left to Right) Original image, input image, generated edges, inpainted results without any post-processing.}
		\label{fig:results}
	\end{figure}

	Figure \ref{fig:results} shows a sample of images generated by our model. For visualization purposes, we reverse the colors of $\mathbf{C}_{comp}$. Our model is able to generate photo-realistic results with a large fraction of image structures remaining intact. Furthermore, by including style loss, the inpainted images lack any ``checkerboard'' artifacts in the generated results. As importantly, the inpainted images exhibit minimal blurriness. 
	Figure \ref{fig:compare} compares images generated by our method with those generated by other state-of-the-art techniques. The images generated by our proposed model are closer to ground truth than images from other methods. We conjecture that when edge information is present, the network only needs to learn the color distribution, without having to worry about preserving image structure.
	
	\subsection{Quantitative Comparison}
	\paragraph{Numerical Metrics} We measure the quality of our results using the following metrics: 1) relative $\ell_1$; 2) structural similarity index (SSIM) \cite{wang2004image}, with a window size of $11$; and 3) peak signal-to-noise ratio (PSNR). These metrics assume pixel-wise independence, which may assign favorable scores to perceptually inaccurate results. Therefore, we also include Fr\'{e}chet Inception Distance (FID) \cite{heusel2017gans}. Recent works \cite{zhang2018unreasonable, zhang2018self, dolhansky2018eye} have shown that metrics based on deep features are closer to those based on human perception. FID measures the Wasserstein-2 distance between the feature space representations of real and inpainted images using a pre-trained Inception-V3 model \cite{szegedy2016rethinking}. The results over Places2 dataset are reported in Table \ref{tab:places}. Note that these statistics are based on the synthesized image which mostly comprises of the ground truth image. Therefore our reported FID values are lower than other generative models reported in \cite{lucic2017gans}.
	\begin{table}[h!]
		\def\arraystretch{1.02}
		\centering
		\begin{tabular}{c|c*{5}{|>{\centering\arraybackslash}p{.09\linewidth}}}
			\multicolumn{2}{r|}{\textbf{Mask}} & \small{CA} & \small{GLCIC} & \small{PConv*} & \small{Ours} & \small{Canny} \\ \hhline{*{6}{=|}=}
			\multirow{ 5}{*}{\rotatebox{90}{$\ell_1$ (\%)$^{\dagger}$}}
			& \small{10-20\%} & 2.41 & 2.66 & \textbf{1.14} & 1.50 & 1.16 \\ \cline{2-7}
			& \small{20-30\%} & 4.23 & 4.70 & \textbf{1.98} & 2.59 & 1.88 \\ \cline{2-7}
			& \small{30-40\%} & 6.15 & 6.78 & \textbf{3.02} & 3.77 & 2.60 \\ \cline{2-7}
			& \small{40-50\%} & 8.03 & 8.85 & \textbf{4.11} & 5.14 & 3.41 \\ \cline{2-7}
			& \small{Fixed} & 4.37 & 4.12 & - & \textbf{3.86} & 2.22 \\ \hhline{*{6}{=|}=}
			\multirow{ 5}{*}{\rotatebox{90}{SSIM$^{\star}$}}
			& \small{10-20\%} & 0.893 & 0.862 & 0.869 & \textbf{0.920} & 0.941 \\ \cline{2-7}
			& \small{20-30\%} & 0.815 & 0.771 & 0.777 & \textbf{0.861} & 0.902 \\ \cline{2-7}
			& \small{30-40\%} & 0.739 & 0.686 & 0.685 & \textbf{0.799} & 0.863 \\ \cline{2-7}
			& \small{40-50\%} & 0.662 & 0.603 & 0.589 & \textbf{0.731} & 0.821 \\ \cline{2-7}
			& \small{Fixed} & 0.818 & 0.814 & - & \textbf{0.823} & 0.892 \\ \hhline{*{6}{=|}=}
			\multirow{ 5}{*}{\rotatebox{90}{PSNR$^{\star}$}}
			& \small{10-20\%} & 24.36 & 23.49 & \textbf{28.02} & 27.95 & 30.85 \\ \cline{2-7}
			& \small{20-30\%} & 21.19 & 20.45 & 24.90 & \textbf{24.92} & 28.35 \\ \cline{2-7}
			& \small{30-40\%} & 19.13 & 18.50 & 22.45 & \textbf{22.84} & 26.66 \\ \cline{2-7}
			& \small{40-50\%} & 17.75 & 17.17 & 20.86 & \textbf{21.16} & 25.20 \\ \cline{2-7}
			& \small{Fixed} & 20.65 & 21.34 & - & \textbf{21.75} & 26.52 \\ \hhline{*{6}{=|}=}
			\multirow{ 5}{*}{\rotatebox{90}{FID$^{\dagger}$}}
			& \small{10-20\%} & 6.16 & 11.84 & - & \textbf{2.32} & 2.25 \\ \cline{2-7}
			& \small{20-30\%} & 14.17 & 25.11 & - & \textbf{4.91} & 3.42 \\ \cline{2-7}
			& \small{30-40\%} & 24.16 & 39.88 & - & \textbf{8.91} & 4.87 \\ \cline{2-7}
			& \small{40-50\%} & 35.78 & 54.30 & - & \textbf{14.98} & 7.13 \\ \cline{2-7}
			& \small{Fixed} & 8.31 & 8.42 & - & \textbf{8.16} & 3.24 \\ \hline
		\end{tabular}
		\caption{Quantitative results over Places2 with models: Contextual Attention (CA) \cite{yu2018generative}, Globally and Locally Consistent Image Completion (GLCIC) \cite{iizuka2017globally}, Partial Convolution (PConv) \cite{Liu_2018_ECCV}, $G_1$ and $G_2$ (Ours), $G_2$ only with Canny edges (Canny). The best result of each row is boldfaced except for Canny. *Values taken from the paper \cite{Liu_2018_ECCV}. $^\dagger$Lower is better. $^\star$Higher is better.}
		\label{tab:places}
	\end{table}
	\begin{figure}[h!]
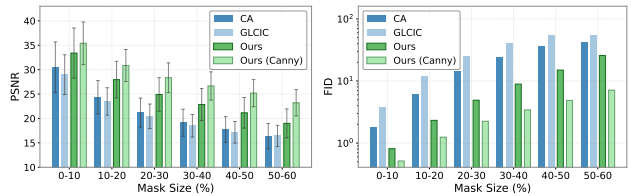

		\centering
		\begin{subfigure}[l]{.235\textwidth}
			\centering
			\includegraphics[width=\textwidth]{barchart_places_psnr.pdf}
		\end{subfigure} 
		\begin{subfigure}[r]{.235\textwidth}
			\centering
			\includegraphics[width=\textwidth]{barchart_places_fid.pdf}
		\end{subfigure}
		\caption{Effect of mask sizes on PSNR and FID. (Places2)}
		\label{fig:mask}
	\end{figure}

	Figure \ref{fig:mask} shows the performance of our model for various mask sizes. Statistics for competing techniques are obtained using their respective pre-trained weights, where available\footnote{\scriptsize \url{https://github.com/JiahuiYu/generative_inpainting}} \footnote{\scriptsize \url{https://github.com/satoshiiizuka/siggraph2017_inpainting}}. Results for Partial Convolution (PConv) \cite{Liu_2018_ECCV} are taken from their paper as the source code is not available at the time of writing. Note that $\ell_1$ (\%) errors for PConv are lower than those achieved by our method and those reported in CA \cite{yu2018generative} and GLCIC \cite{iizuka2017globally}. While we are not sure why this is so, we suspect PConv is computing this score differently than how we compute it. Our statistics are calculated over $10,000$ random images in the test set.
	\paragraph{Visual Turing Tests} We evaluate our results using the human perceptual metrics \emph{2 alternative forced choice} (2AFC) and \emph{just noticeable differences} (JND). For 2AFC, we ask participants whether or not an image is real from a pool of randomized images. For JND, we ask participants to select the more realistic image from pairs of real and generated images. Participants are given two seconds for each test. The tests are performed over 300 images for each model and mask size. Each image is shown 10 times in total. The results are summarized in Table \ref{tab:turk}.
	\begin{table}[h!]
		\centering
		\small
		\def\arraystretch{1.15}
		\begin{tabular}{c|c*{3}{|>{\centering\arraybackslash}p{.2\linewidth}}}
			\multicolumn{2}{r|}{\textbf{Mask}~ } & CA & GLCIC & Ours \\ \hhline{*{4}{=|}=}
			\multirow{ 4}{*}{\rotatebox{90}{JND (\%)}} & 10-20\% & 21 $\pm$ 1.2\% & 16.9 $\pm$ 1.1\% & \textbf{39.7 $\pm$ 1.5\%} \\ \cline{2-5}
			& 20-30\% & 15.5 $\pm$ 1.1\% & 14.3 $\pm$ 1\% & \textbf{37 $\pm$ 1.5\%} \\ \cline{2-5}
			& 30-40\% & 12.9 $\pm$ 1\% & 12.3 $\pm$ 1\% & \textbf{27.5 $\pm$ 1.3\%} \\ \cline{2-5}
			& 40-50\% & 12.7 $\pm$ 2\% & 10.9 $\pm$ 0.9\% & \textbf{25.4 $\pm$ 1.3\%} \\ \hhline{*{4}{=|}=}
			\multirow{ 4}{*}{\rotatebox{90}{2AFC (\%)}} & 10-20\% & 38.7 $\pm$ 1.8\% & 22.5 $\pm$ 1.5\% & \textbf{88.7 $\pm$ 1.2\% } \\ \cline{2-5}
			& 20-30\% & 23.4 $\pm$ 1.5\% & 12.1 $\pm$ 1.2\% & \textbf{77.6 $\pm$ 1.5\%} \\ \cline{2-5}
			& 30-40\% & 13.5 $\pm$ 1.3\% & 4.3 $\pm$ 0.7\% & \textbf{66.4 $\pm$ 1.8\%} \\ \cline{2-5}
			& 40-50\% & 9.9 $\pm$ 1\% & 2.8 $\pm$ 0.6\% & \textbf{58 $\pm$ 1.8\%} \\ \hline
		\end{tabular}
		\caption{2AFC and JND scores for various mask sizes on Places2. 2AFC score for ground truth is \textbf{94.6 $\pm$ 0.5\%}.}
		\label{tab:turk}
	\end{table}
	\subsection{Ablation Study} \label{ablation}
	\paragraph{Quantity of Edges versus Inpainting Quality} We now turn our attention to the key assumption of this work: edge information helps with image inpainting. Table \ref{tab:compare_sigma} shows inpainting results with and without edge information. Our model achieved better scores for every metric when edge information was incorporated into the inpainting model, even when a significant portion of the image is missing.
	\begin{table}[h!]
		\centering
		\begin{tabular}{c*{4}{|>{\centering\arraybackslash}p{.13\linewidth}}}
			~ & \multicolumn{2}{c|}{CelebA} & \multicolumn{2}{c}{Places2} \\ \cline{2-5}
			Edges & No & Yes & No & Yes \\ \hhline{*{4}{=|}=}
			$\ell_1$ (\%) & 4.11 & 3.03 & 6.69 & 5.14 \\ \hline
			SSIM & 0.802 & 0.846 & 0.682 & 0.731 \\ \hline
			PSNR & 23.33 & 25.28 & 19.59 & 21.16 \\ \hline
			FID & 6.16 & 2.82 & 32.18 & 14.98 \\ \hline
		\end{tabular}
		\caption{Comparison of inpainting results with edge information (our full model) and without edge information ($G_2$ only, trained without edges). Statistics are based on $10,000$ random masks with size 40-50\% of the entire image.}
		\label{tab:compare_sigma}
	\end{table}

\begin{figure}[h!]
		\centering
		\begin{subfigure}[l]{.24\textwidth}
			\centering
			\includegraphics[width=\textwidth]{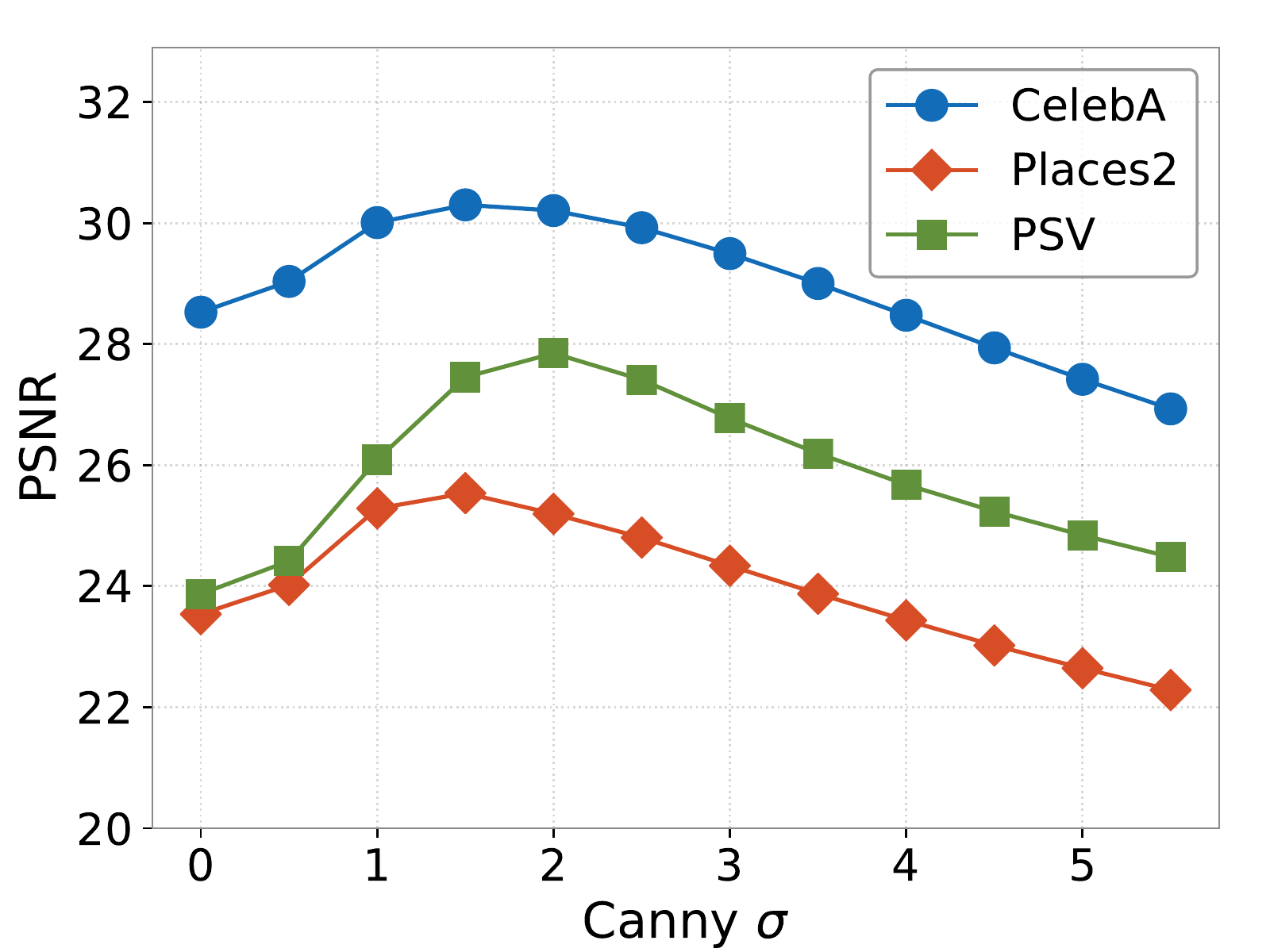}
		\end{subfigure}
		\begin{subfigure}[r]{.24\textwidth}
			\centering
			\includegraphics[width=\textwidth]{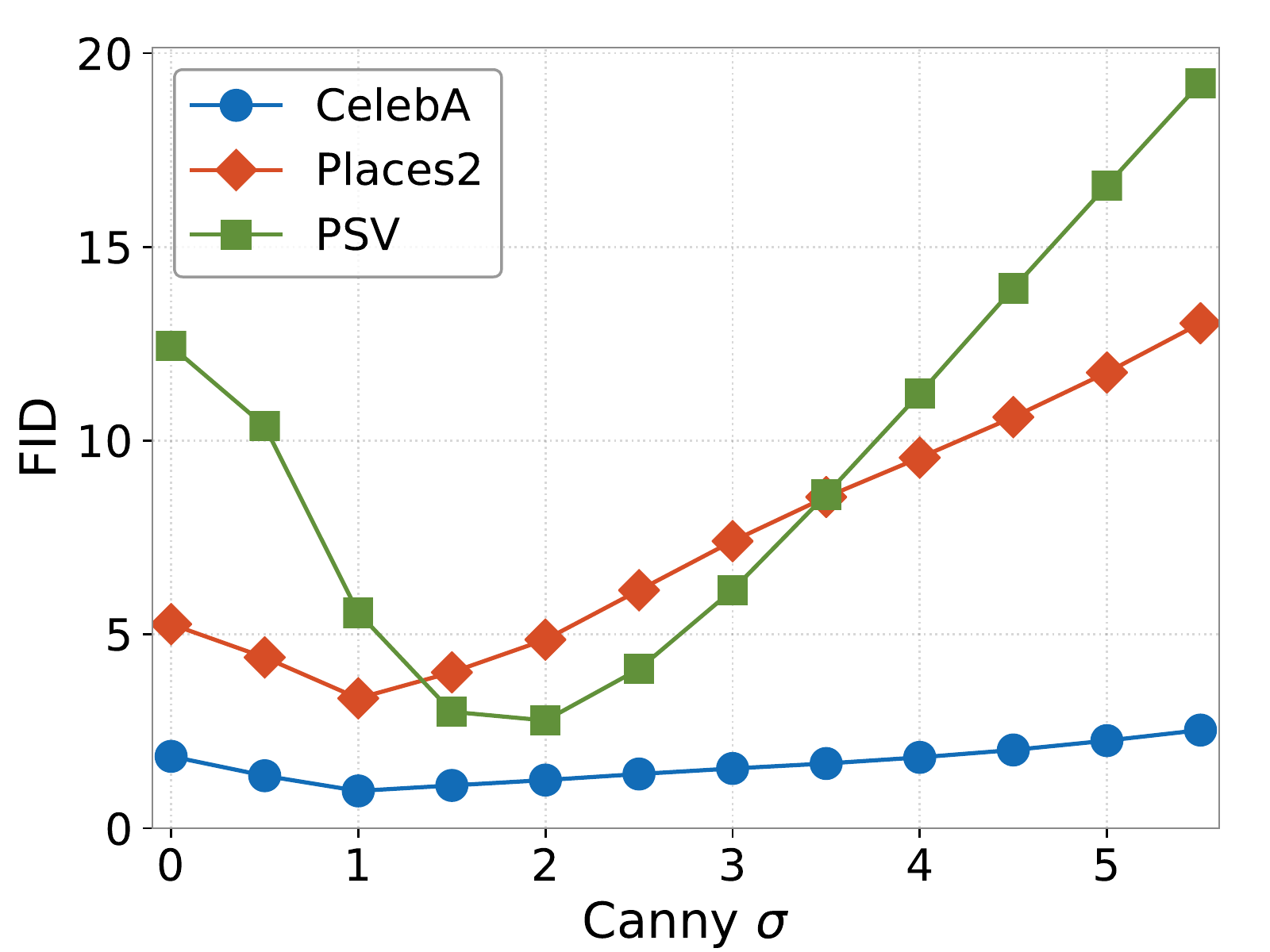}
		\end{subfigure}
		\caption{Effect of $\sigma$ in Canny detector on PSNR and FID.}
		\label{fig:sig}
	\end{figure}

	Next, we turn to a more interesting question: How much edge information is needed to see improvements in the generated images? We again use Canny edge detector to construct edge information. We use the parameter $\sigma$ to control the amount of edge information available to the image completion network. Specifically, we train our image completion network using edge maps generated for $\sigma = 0, 0.5, \dots, 5.5$, and we found that the best image inpainting results are obtained with edges corresponding to $\sigma \in [1.5, 2.5]$, across all datasets shown in Figure \ref{fig:sig}. For large values of $\sigma$, too few edges are available to make a difference in the quality of generated images. On the other hand, when $\sigma$ is too small, too many edges are produced, which adversely affect the quality of the generated images. We used this study to set $\sigma = 2$ when creating ground truth edge maps for the training of the edge generator network.

\begin{figure}[h!]
		\centering
		\includegraphics[width=.4\textwidth]{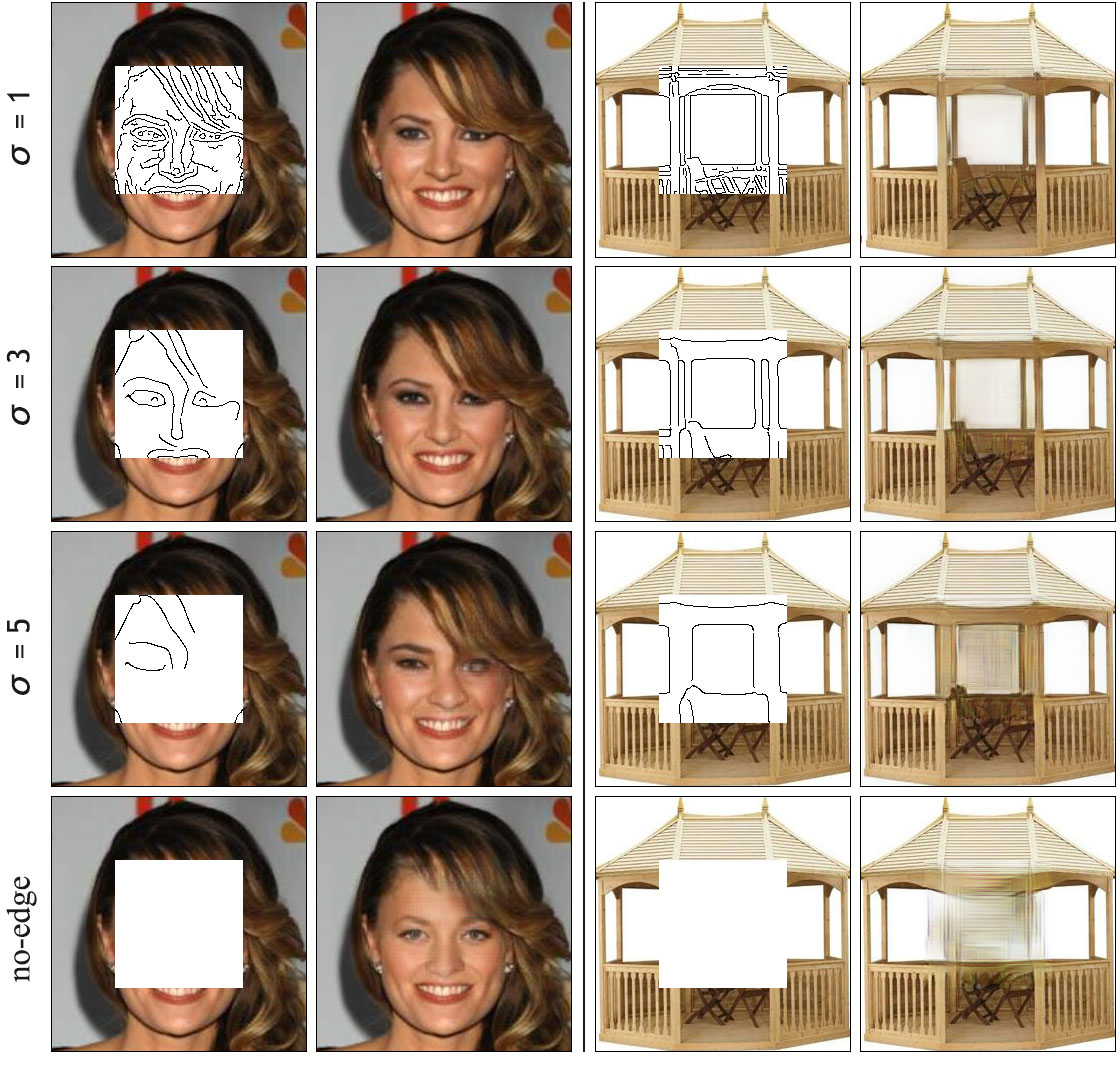}
		\caption{Effect of $\sigma$ in Canny edge detector on inpainting results. Top to bottom: $\sigma = 1, 3, 5$, no edge data.}
		\label{fig:canny}
	\end{figure}

\begin{figure}[h]
		\centering
		\begin{subfigure}[c]{.38\textwidth}
			\centering
			\includegraphics[width=\textwidth]{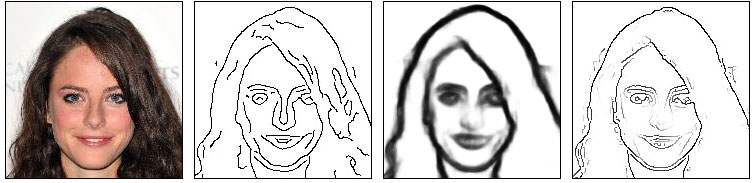}
			\vspace{-15px}
			\caption*{(a)$\qquad\qquad$(b)$\qquad\qquad$(c)$\qquad\qquad$(d)}
		\end{subfigure}
		\vspace{-5px}
		\caption{(a) Image. (b) Canny. (c) HED. (d) Canny$\odot$HED.}
		\label{fig:edges}
	\end{figure}

	Figure \ref{fig:canny} shows how different values of $\sigma$ affects the inpainting task. Note that in a region where edge data is sparse, the quality of the inpainted region degrades. For instance, in the generated image for $\sigma = 5$, the left eye was reconstructed much sharper than the right eye.
	\paragraph{Alternative Edge Detection Systems} We use Canny edge detector to produce training labels for the edge generator network due to its speed, robustness, and ease of use. Canny edges are one-pixel wide, and are represented as binary masks (1 for edge, 0 for background). Edges produced with HED \cite{xie2015holistically}, however, are of varying thickness, and pixels can have intensities ranging between 0 and 1. We noticed that it is possible to create edge maps that look eerily similar to human sketches by performing element-wise multiplication on Canny and HED edge maps (Figure \ref{fig:edges}). We trained our image completion network using the combined edge map. However, we did not notice any improvements in the inpainting results.\footnote{Further analysis with HED are available in appendix \ref{app:hed}.}

\section{Discussions and Future Work}

	We proposed EdgeConnect, a new deep learning model for image inpainting tasks. EdgeConnect comprises of an edge generator and an image completion network, both following an adversarial model. We demonstrate that edge information plays an important role in the task of image inpainting. Our method achieves state-of-the-art results on standard benchmarks, and is able to deal with images with multiple, irregularly shaped missing regions.
	
\begin{figure}[h]
		\centering
		\begin{subfigure}[c]{.11\textwidth}
			\centering
			\includegraphics[width=\textwidth]{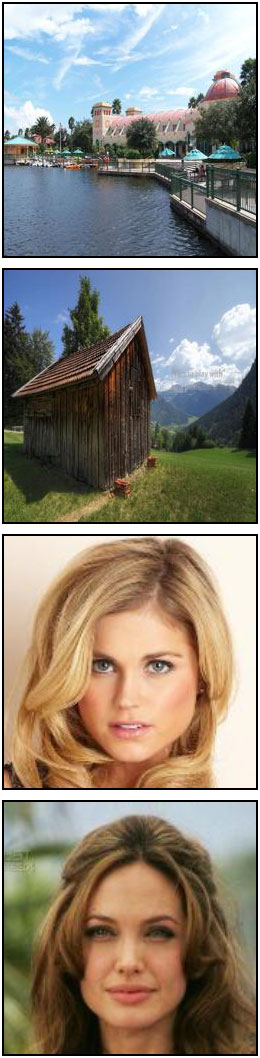}
			\caption{~}
		\end{subfigure} 
		\begin{subfigure}[c]{.11\textwidth}
			\centering
			\includegraphics[width=\textwidth]{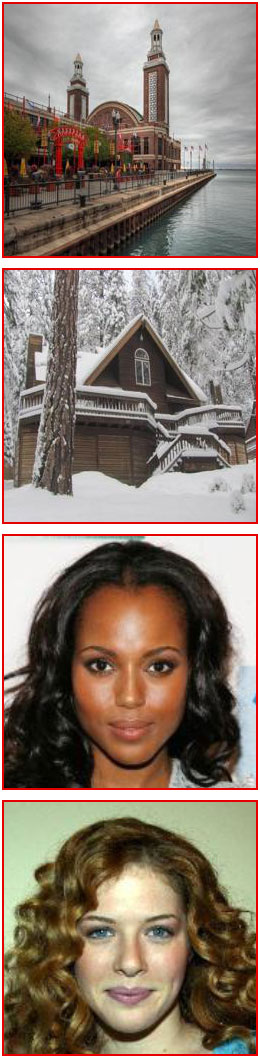}
			\caption{~}
		\end{subfigure}
		\begin{subfigure}[c]{.11\textwidth}
			\centering
			\includegraphics[width=\textwidth]{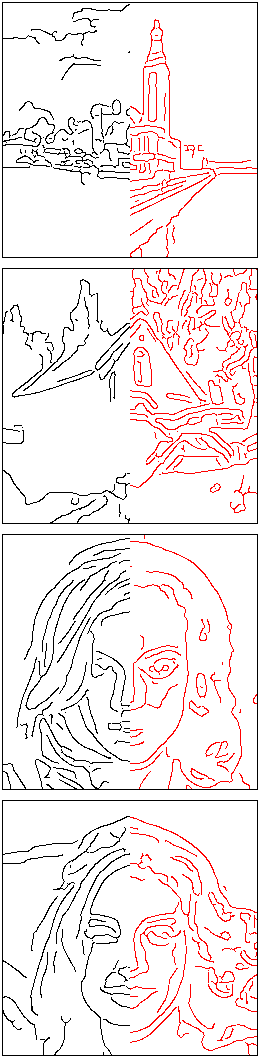}
			\caption{~}
		\end{subfigure} 
		\begin{subfigure}[c]{.11\textwidth}
			\centering
			\includegraphics[width=\textwidth]{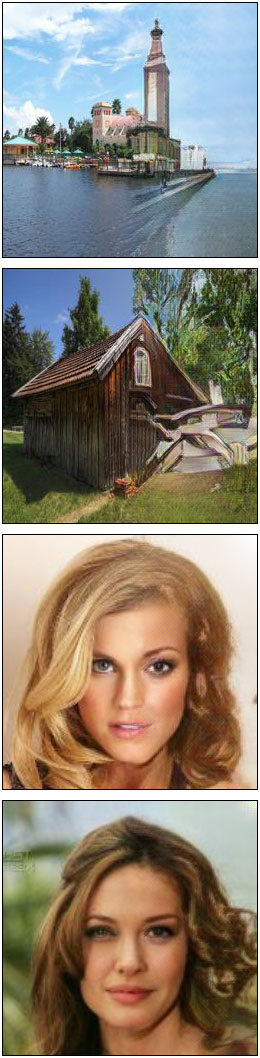}
			\caption{~}
		\end{subfigure} 
		\caption{Edge-map (c) generated using the left-half of (a) (shown in black) and right-half of (b) (shown in red). Input is (a) with the right-half removed, producing the output (d).}
		\label{fig:merge}
	\end{figure}

\begin{figure}[h]
		\centering
		\includegraphics[width=.39\textwidth]{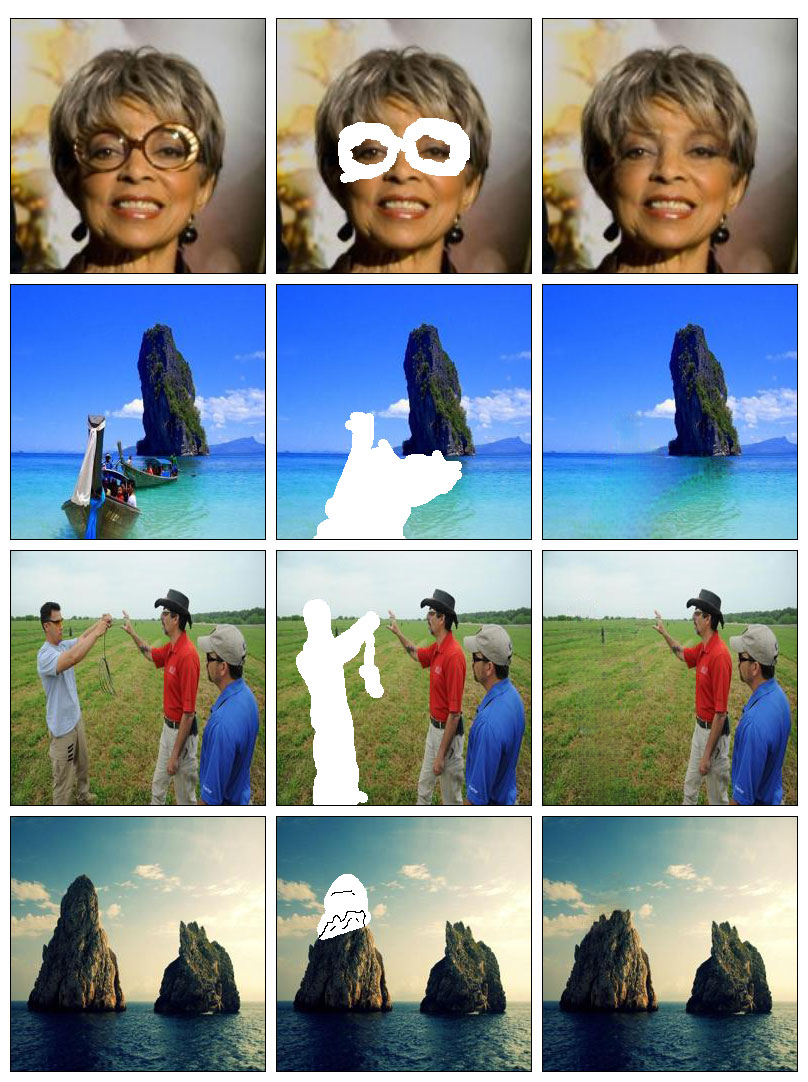}
		\caption{Examples of object removal and image editing using our EdgeConnect model. (Left) Original image. (Center) Unwanted object removed with optional edge information to guide inpainting. (Right) Generated image.}
		\label{fig:remove}
	\end{figure}

	The trained model can be used as an interactive image editing tool. We can, for example, manipulate objects in the edge domain and transform the edge maps back to generate a new image. This is demonstrated in Figure \ref{fig:merge}. Here we have removed the right-half of a given image to be used as input. The edge maps, however, are provided by a different image. The generated image seems to share characteristics of the two images. Figure \ref{fig:remove} shows examples where we attempt to remove unwanted objects from existing images.
	
\begin{figure}[!h]
		\centering
		\includegraphics[width=.36\textwidth]{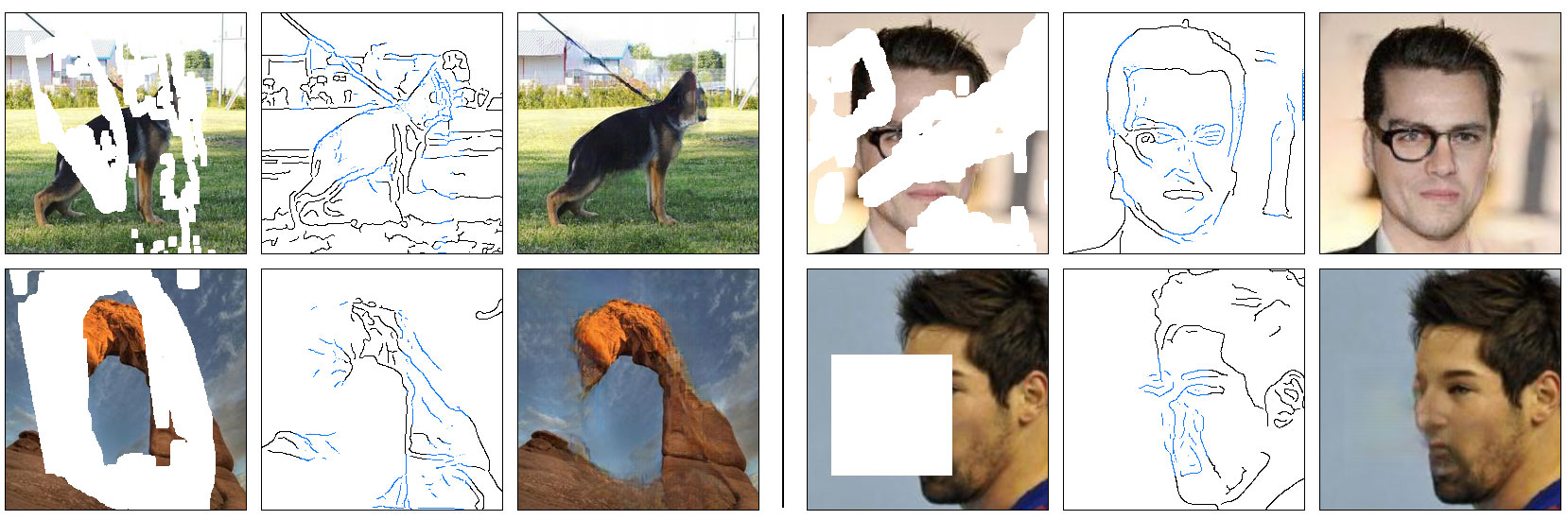}
		\caption{Inpainted results where edge generator fails to produce relevant edge information.}
		\label{fig:fail}
	\end{figure}

	We plan to investigate better edge detectors. While effectively delineating the edges is more useful than hundreds of detailed lines, our edge generating model sometimes fails to accurately depict the edges in highly textured areas, or when a large portion of the image is missing (as seen in Figure \ref{fig:fail}). We believe our fully convolutional generative model can be extended to very high-resolution inpainting applications with an improved edge generating system.

	\newpage
	{
		\footnotesize
		\frenchspacing
		\paragraph{Acknowledgments:}
		This research was supported in part by NSERC Discovery Grant. The authors would like to thank Konstantinos G. Derpanis for helpful discussions and feedbacks. We gratefully acknowledge the support of NVIDIA Corporation with the donation of the Titan V GPU used for this research.
	}

{\small
\bibliographystyle{ieee}
\bibliography{egbib}
}
\newpage
\appendix
\graphicspath{{appendix/}}

\section{Network Architectures} \label{app:netwrok}

	\subsection{Generators} 
	We follow a similar naming convention as those presented in \cite{zhu2017unpaired}. Let \texttt{ck} denote a $7 \times 7$ Convolution-SpectralNorm-InstanceNorm-ReLU layer with $k$ filters and stride $1$ with reflection padding. Let \texttt{dk} denote a $4 \times 4$ Convolution-SpectralNorm-InstanceNorm-ReLU layer with $k$ filters and stride $2$ for down-sampling. Let \texttt{uk} be defined in the same manner as \texttt{dk} with transpose convolution for up-sampling. Let \texttt{Rk} denote a residual block of channel size $k$ across both layers. We use dilated convolution in the first layer of \texttt{Rk} with dilation factor of $2$, followed by spectral normalization and instance normalization.
	
	The architecture of our generators is adopted from the model proposed by Johnson \emph{et al.} \cite{johnson2016perceptual}: \\
	\texttt{c64, d128, d256, R256, R256, R256, R256, R256, R256, R256, R256, u128, u64, c*}.
	
	The final layer \texttt{c*} varies depending on the generator. In the edge generator $G_1$, \texttt{c*} has channel size of $1$ with sigmoid activation for edge prediction. In the image completion network $G_2$, \texttt{c*} has channel size of $3$ with $\tanh$ (scaled) activation for the prediction of RGB pixel intensities. In addition, we remove spectral normalization from all layers of $G_2$.
	
	\subsection{Discriminators}
	The discriminators $D_1$ and $D_2$ follow the same architecture based on the $70 \times 70$ PatchGAN \cite{isola2017image, zhu2017unpaired}. Let \texttt{Ck-s} denote a $4 \times 4$ Convolution-SpectralNorm-LeakyReLU layer with $k$ filters of stride $s$. The discriminators have the architecture \texttt{C64-2, C128-2, C256-2, C512-1, C1-1}. The final convolution layer produces scores predicting whether $70 \times 70$ overlapping image patches	are real or fake. LeakyReLU \cite{maas2013rectifier} is employed with slope $0.2$.

\section{Experimental Results} \label{app:results}

	We provide additional results produced by our model over the following datasets:
	\begin{itemize}
		\item CelebA ($202,599$ images)
		\item Places2 ($10$ million+ images)
		\item Paris StreetView ($14,900$ images)
	\end{itemize}
	With CelebA, we cropped the center $178 \times 178$ of the images, then resized them to $256 \times 256$ using bilinear interpolation. For Paris StreetView, since the images in the dataset are elongated ($936 \times 537$), we separate each image into three: 1) Left $537 \times 537$, 2) middle $537 \times 537$, 3) right $537 \times 537$, of the image. These images are scaled down to $256 \times 256$ for our model, totaling $44,700$ images.

	\subsection{Accuracy of Edge Generator}
	Table \ref{tab:precrec} shows the accuracy of our edge generator $G_1$ across all three datasets. We measure precision and recall for various mask sizes. We emphasize that the goal of this experiment is not to achieve best precision and recall results, but instead to showcase how close the generated edges are to the oracle.
	\begin{table}[htb!]
	\centering
	\def\arraystretch{1.2}
	\begin{tabular}{c|c*{2}{|>{\centering\arraybackslash}p{.21\linewidth}}}
		\multicolumn{2}{r|}{\textbf{Mask}} & Precision & Recall \\
		\hhline{=*{2}{=|}=}
		\multirow{6}{*}{\rotatebox{90}{CelebA}}
		& \small{0-10\%} & 40.24 & 38.23 \\ \cline{2-4}
		& \small{10-20\%} & 34.28 & 34.05 \\ \cline{2-4}
		& \small{20-30\%} & 29.23 & 30.07 \\ \cline{2-4}
		& \small{30-40\%} & 24.92 & 25.88 \\ \cline{2-4}
		& \small{40-50\%} & 21.73 & 22.55 \\ \cline{2-4}
		& \small{50-60\%} & 16.39 & 16.93 \\ \hhline{*{3}{=|}=}
		
		\multirow{6}{*}{\rotatebox{90}{Places2}}
		& \small{0-10\%} & 37.87 & 36.41 \\ \cline{2-4}
		& \small{10-20\%} & 32.66 & 32.53 \\ \cline{2-4}
		& \small{20-30\%} & 28.36 & 28.92 \\ \cline{2-4}
		& \small{30-40\%} & 25.02 & 25.38 \\ \cline{2-4}
		& \small{40-50\%} & 22.48 & 22.48 \\ \cline{2-4}
		& \small{50-60\%} & 18.14 & 17.76 \\ \hhline{*{3}{=|}=}
		
		\multirow{6}{*}{\rotatebox{90}{PSV}}
		& \small{0-10\%} & 49.46 & 46.76 \\ \cline{2-4}
		& \small{10-20\%} & 44.31 & 42.31 \\ \cline{2-4}
		& \small{20-30\%} & 39.38 & 37.80 \\ \cline{2-4}
		& \small{30-40\%} & 35.11 & 33.54 \\ \cline{2-4}
		& \small{40-50\%} & 31.29 & 29.60 \\ \cline{2-4}
		& \small{50-60\%} & 24.61 & 22.59 \\ \cline{1-4}
	\end{tabular}
	\caption{Quantitative performance of our edge generator $G_1$ trained on Canny edges.}
	\label{tab:precrec}
	\end{table}
	\newline

	\subsection{Comprehensive Results}
\begin{figure*}[h!]
		\centering
		\begin{subfigure}[l]{.48\linewidth}
			\centering
			\includegraphics[width=\linewidth]{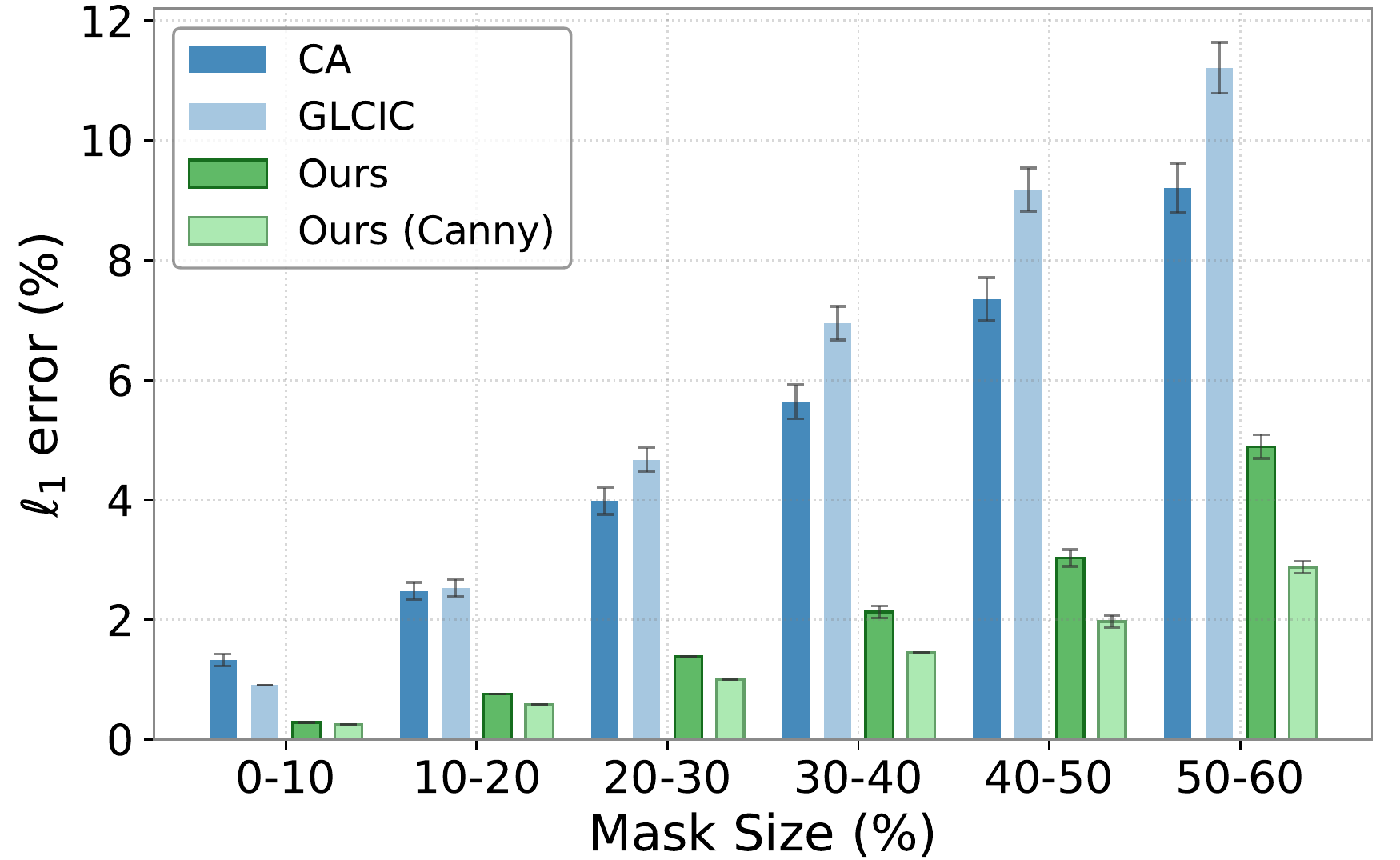}
		\end{subfigure}
		\begin{subfigure}[r]{.48\linewidth}
			\centering
			\includegraphics[width=\linewidth]{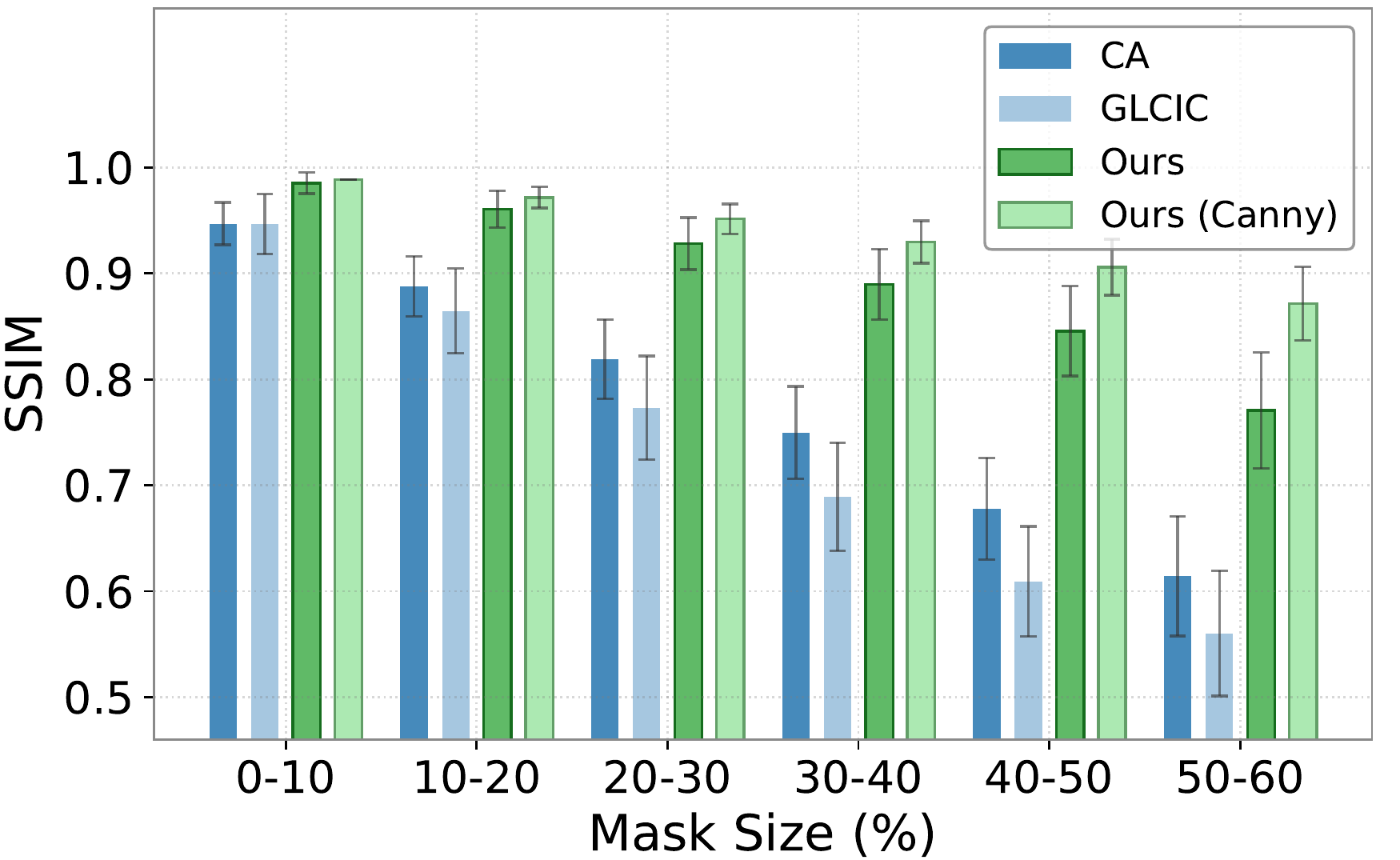}
		\end{subfigure} \\
		\begin{subfigure}[l]{.48\linewidth}
			\centering
			\includegraphics[width=\linewidth]{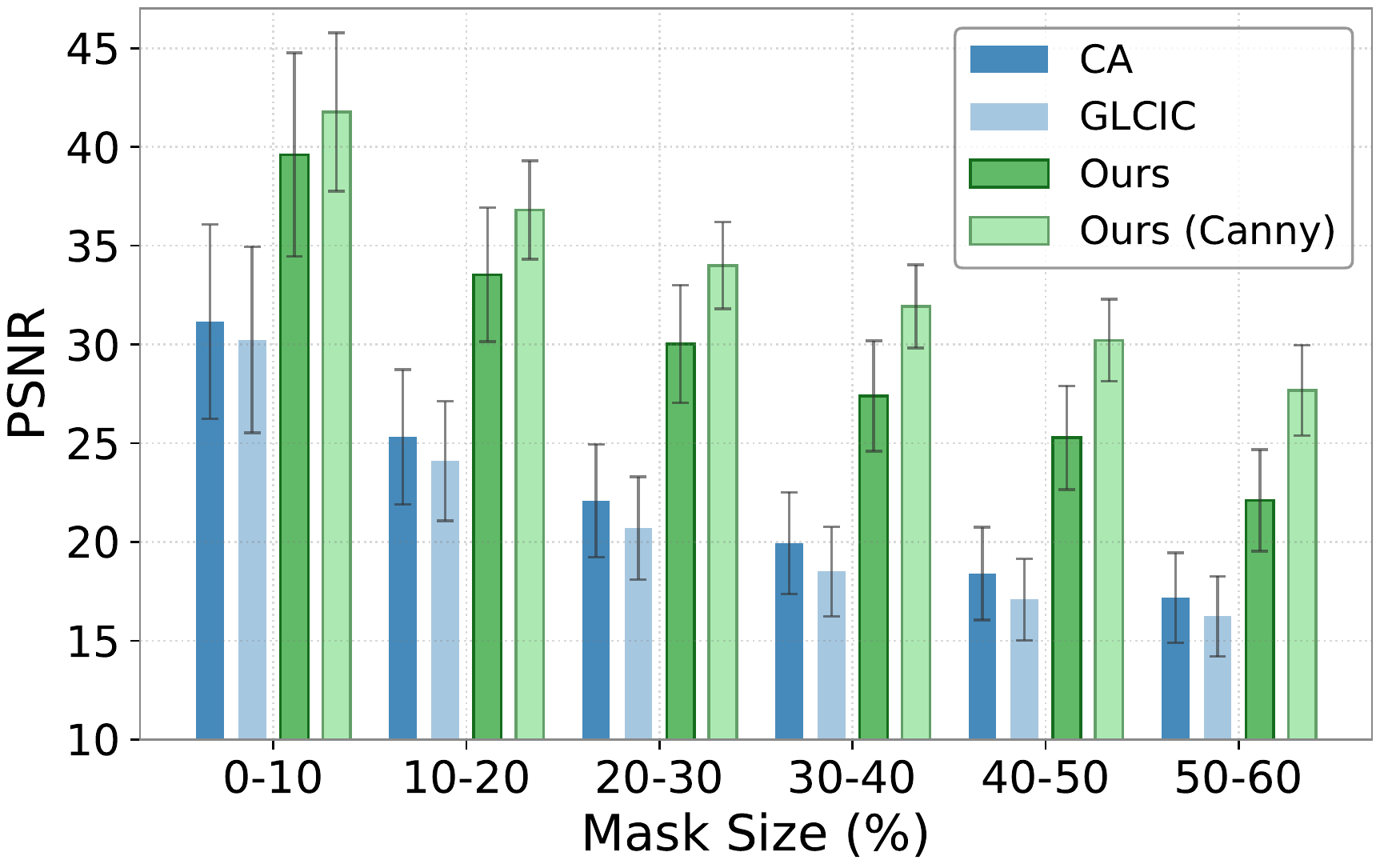}
		\end{subfigure}
		\begin{subfigure}[r]{.48\linewidth}
			\centering
			\includegraphics[width=\linewidth]{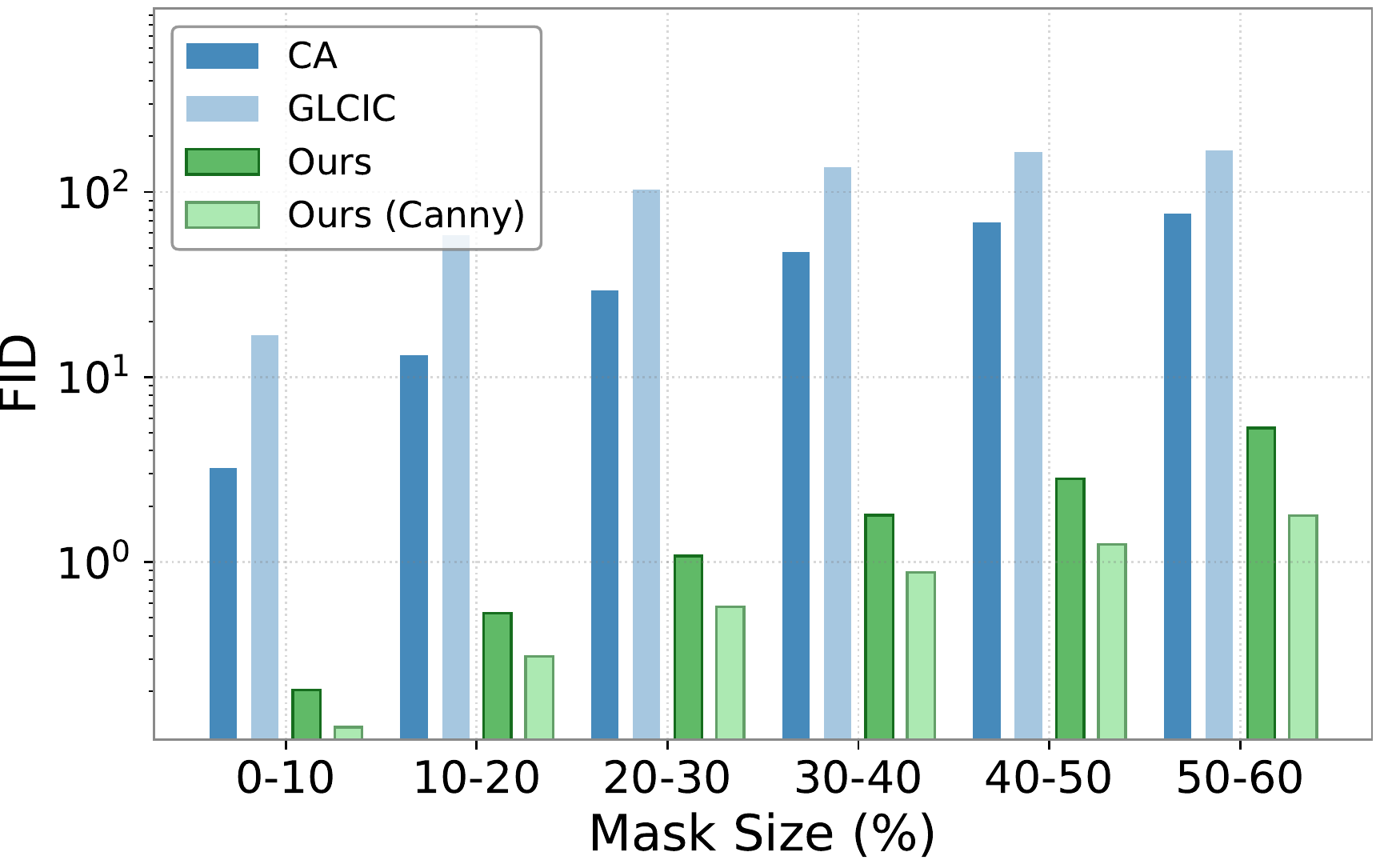}
		\end{subfigure}
		\caption{Effect of relative mask sizes on $\ell_1$, SSIM, PSNR, and FID on the CelebA dataset.}
		\label{fig:bar_celeb}
	\end{figure*}

\begin{figure*}[h!]
		\centering
		\begin{subfigure}[l]{.48\linewidth}
			\centering
			\includegraphics[width=\linewidth]{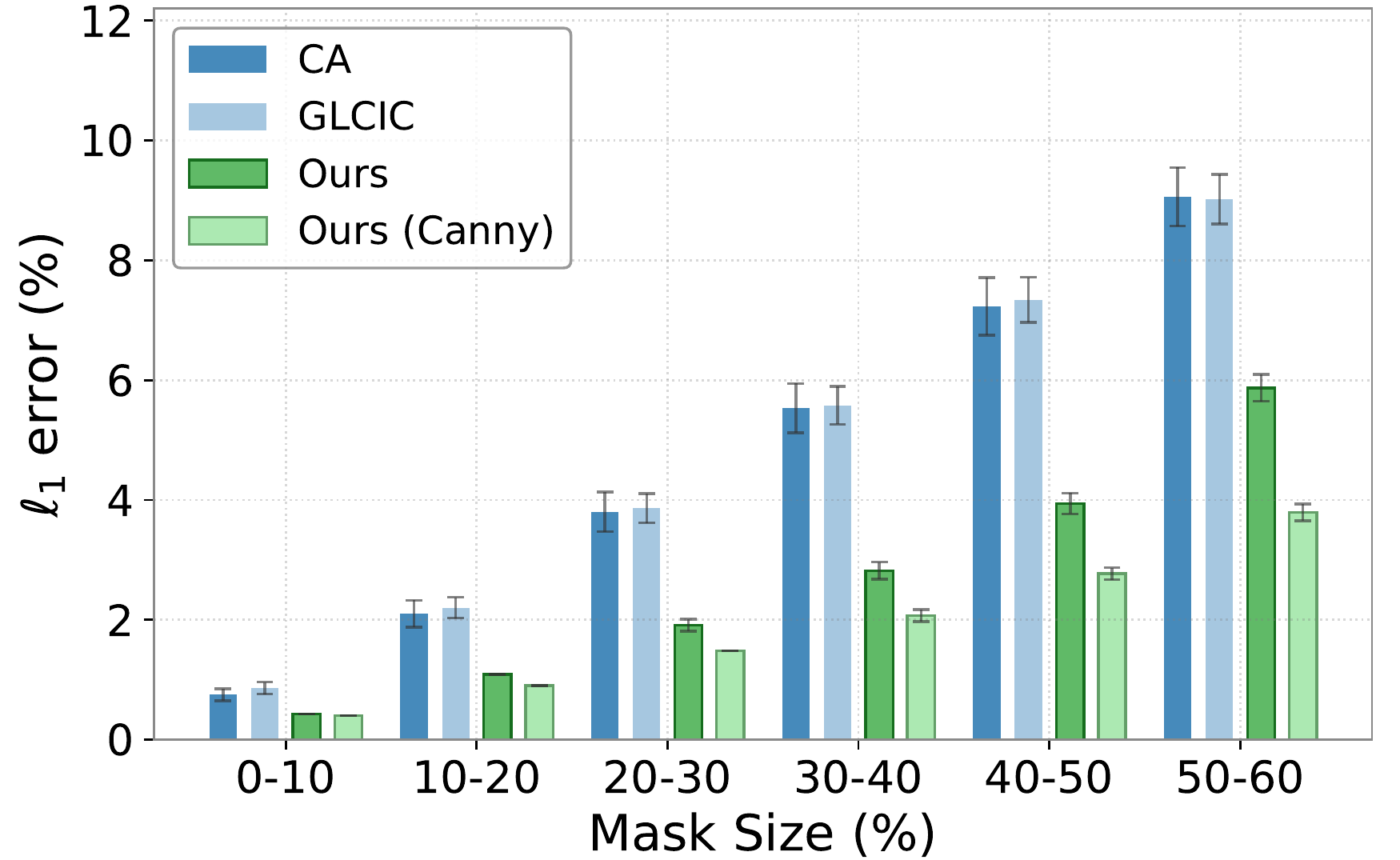}
		\end{subfigure}
		\begin{subfigure}[r]{.48\linewidth}
			\centering
			\includegraphics[width=\linewidth]{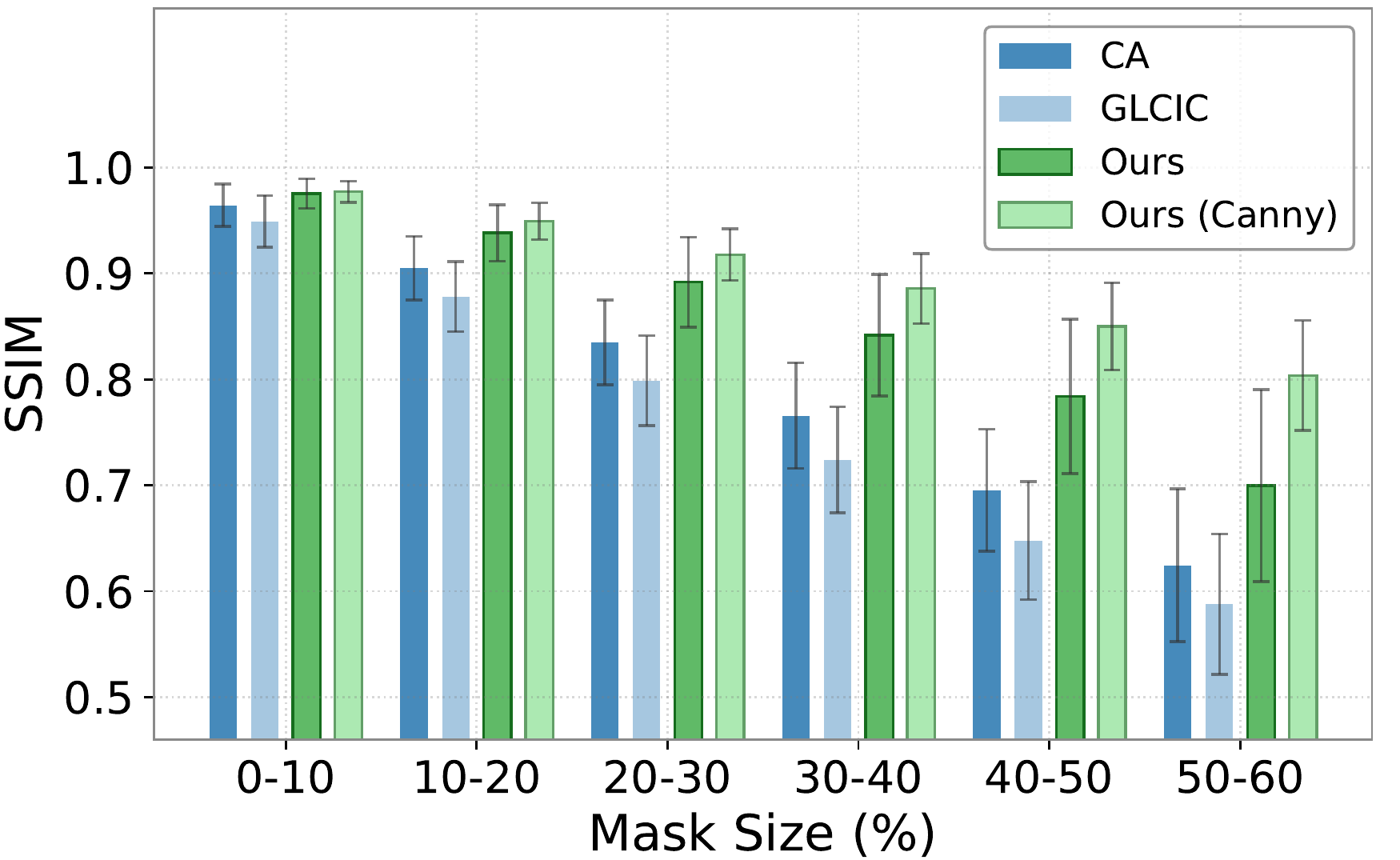}
		\end{subfigure} \\
		\begin{subfigure}[l]{.48\linewidth}
			\centering
			\includegraphics[width=\linewidth]{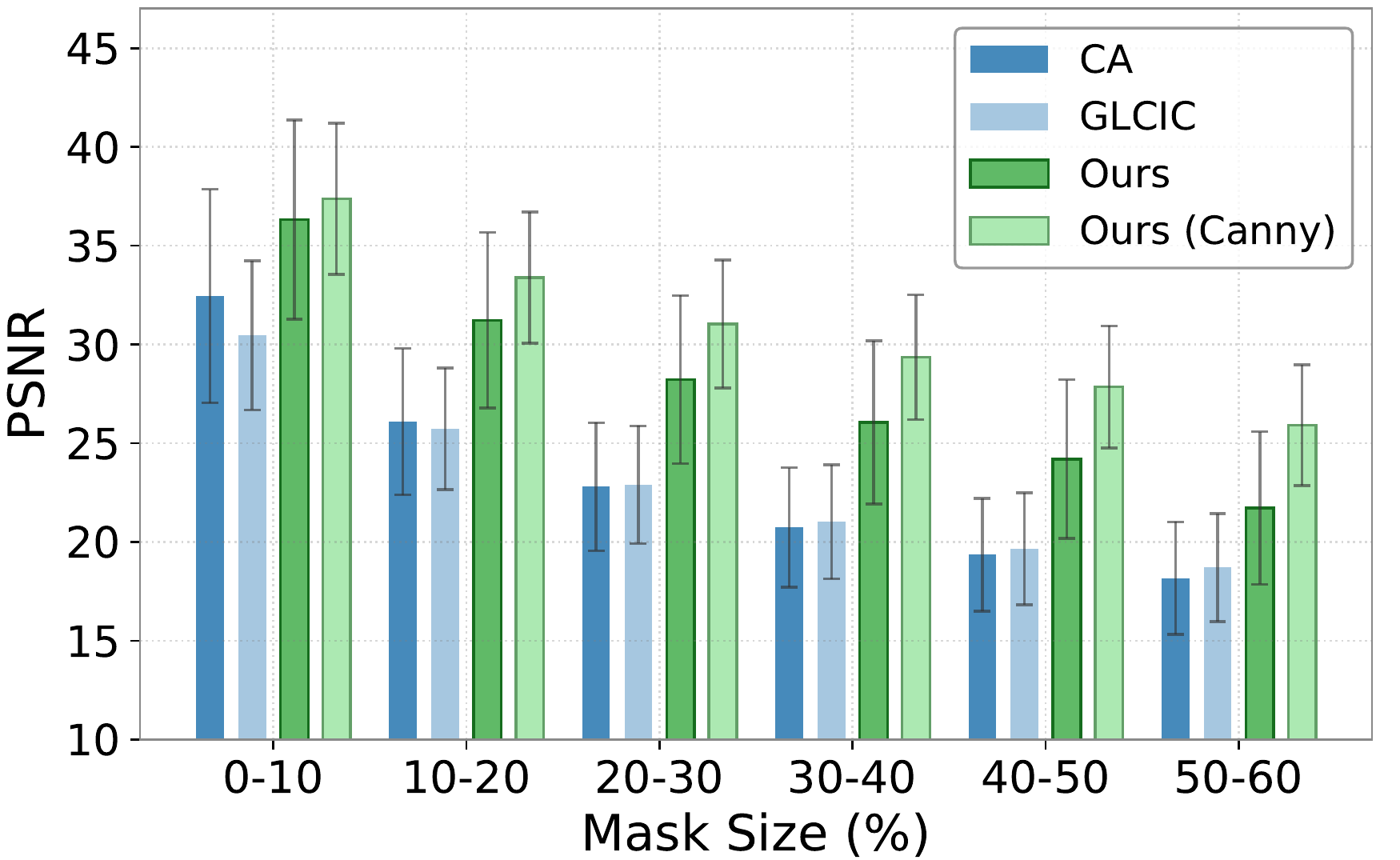}
		\end{subfigure}
		\begin{subfigure}[r]{.48\linewidth}
			\centering
			\includegraphics[width=\linewidth]{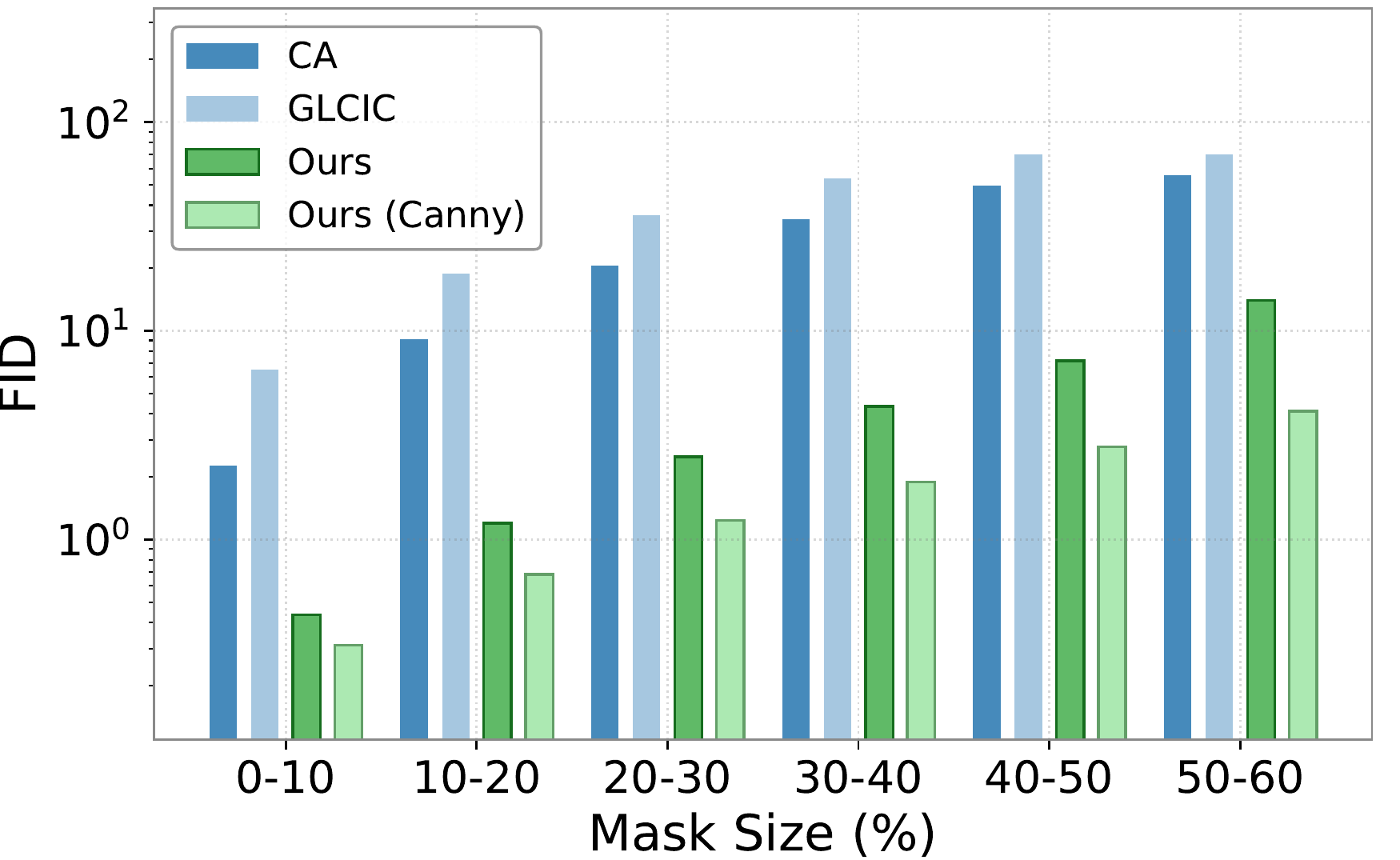}
		\end{subfigure}
		\caption{Effect of relative mask sizes on $\ell_1$, SSIM, PSNR, and FID on the Paris StreetView dataset.}
		\label{fig:bar_psv}
	\end{figure*}

\begin{figure*}[h!]
		\centering
		\begin{subfigure}[c]{\linewidth}
			\centering
			\includegraphics[width=\linewidth]{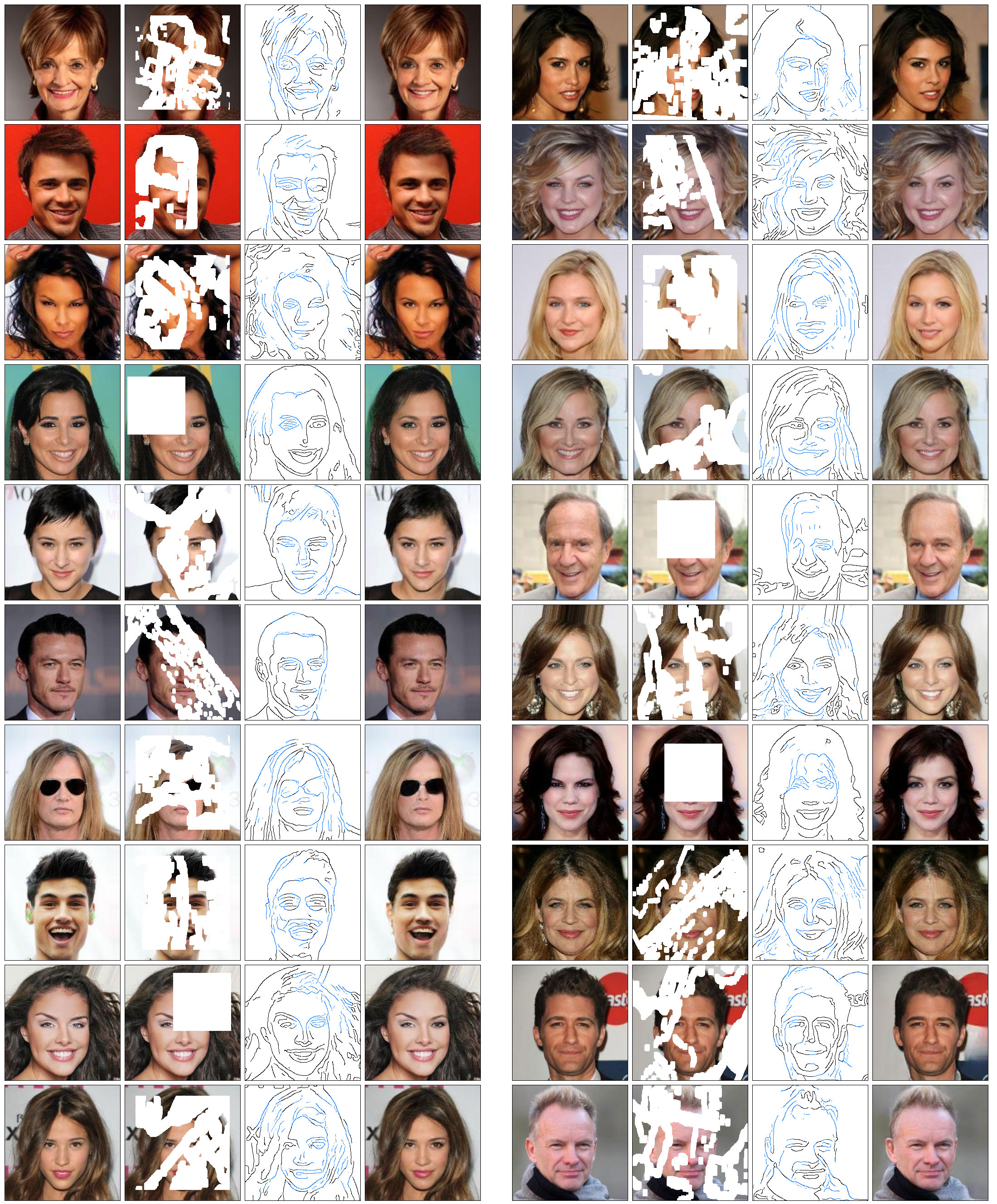}
			\caption*{(a)\hspace{50pt}(b)\hspace{50pt}(c)\hspace{50pt}(d)\hspace{65pt}(a)\hspace{50pt}(b)\hspace{50pt}(c)\hspace{50pt}(d)}
		\end{subfigure}
		\caption{Sample of results with CelebA dataset. Images are best viewed in color. (a) Original Image. (b) Input Image. (c) Generated Edges (Blue) and Ground Truth Edges (Black). (d) Generated Result.}
		\label{fig:celeb}
	\end{figure*}

\begin{figure*}[h!]
		\centering
		\begin{subfigure}[c]{\linewidth}
			\centering
			\includegraphics[width=\linewidth]{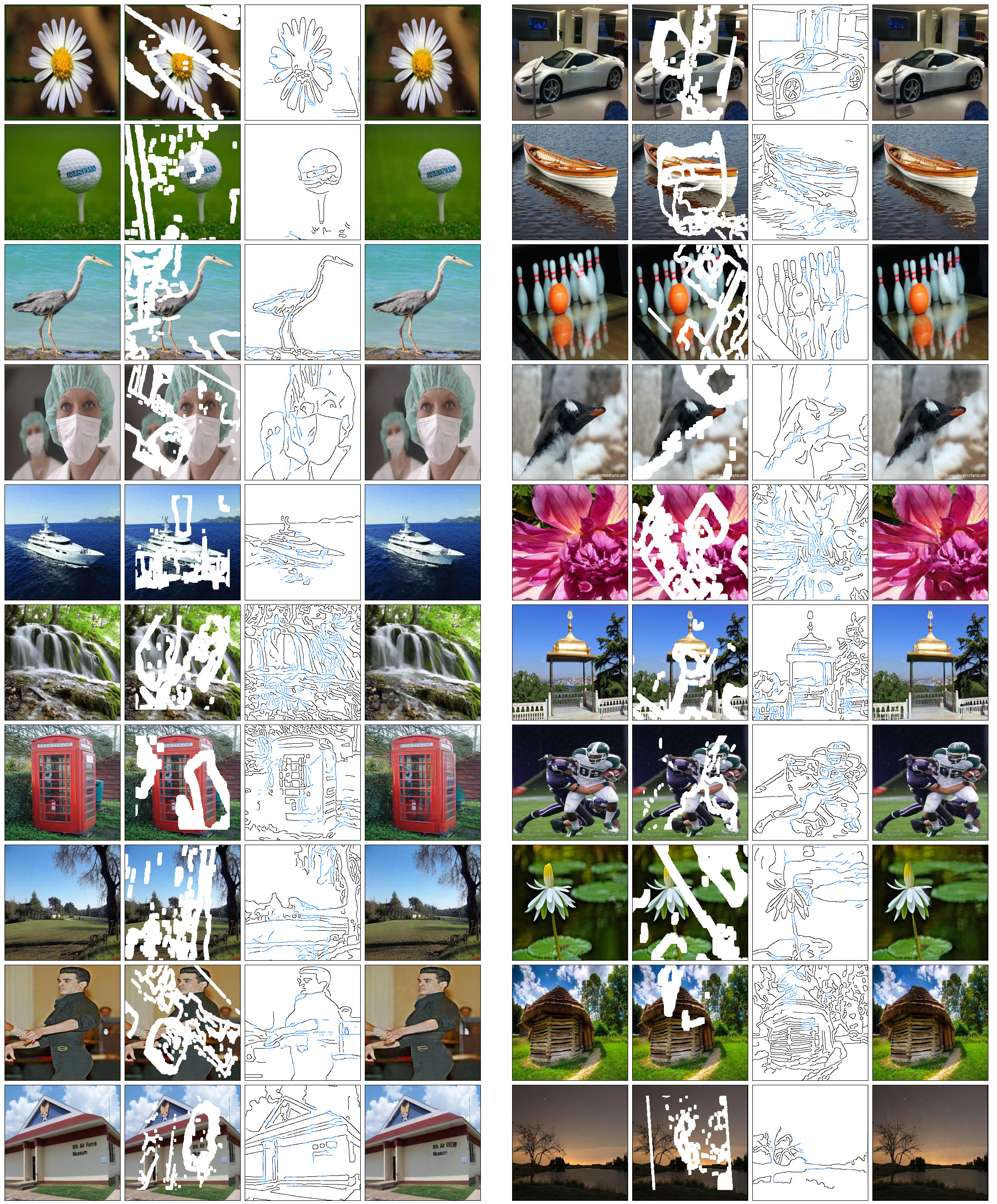}
			\caption*{(a)\hspace{50pt}(b)\hspace{50pt}(c)\hspace{50pt}(d)\hspace{65pt}(a)\hspace{50pt}(b)\hspace{50pt}(c)\hspace{50pt}(d)}
		\end{subfigure}
		\caption{Sample of results with Places2 dataset. Images are best viewed in color. (a) Original Image. (b) Input Image. (c) Generated Edges (Blue) and Ground Truth Edges (Black). (d) Generated Result.}
		\label{fig:places}
	\end{figure*}

\begin{figure*}[h!]
		\centering
		\begin{subfigure}[c]{\linewidth}
			\centering
			\includegraphics[width=\linewidth]{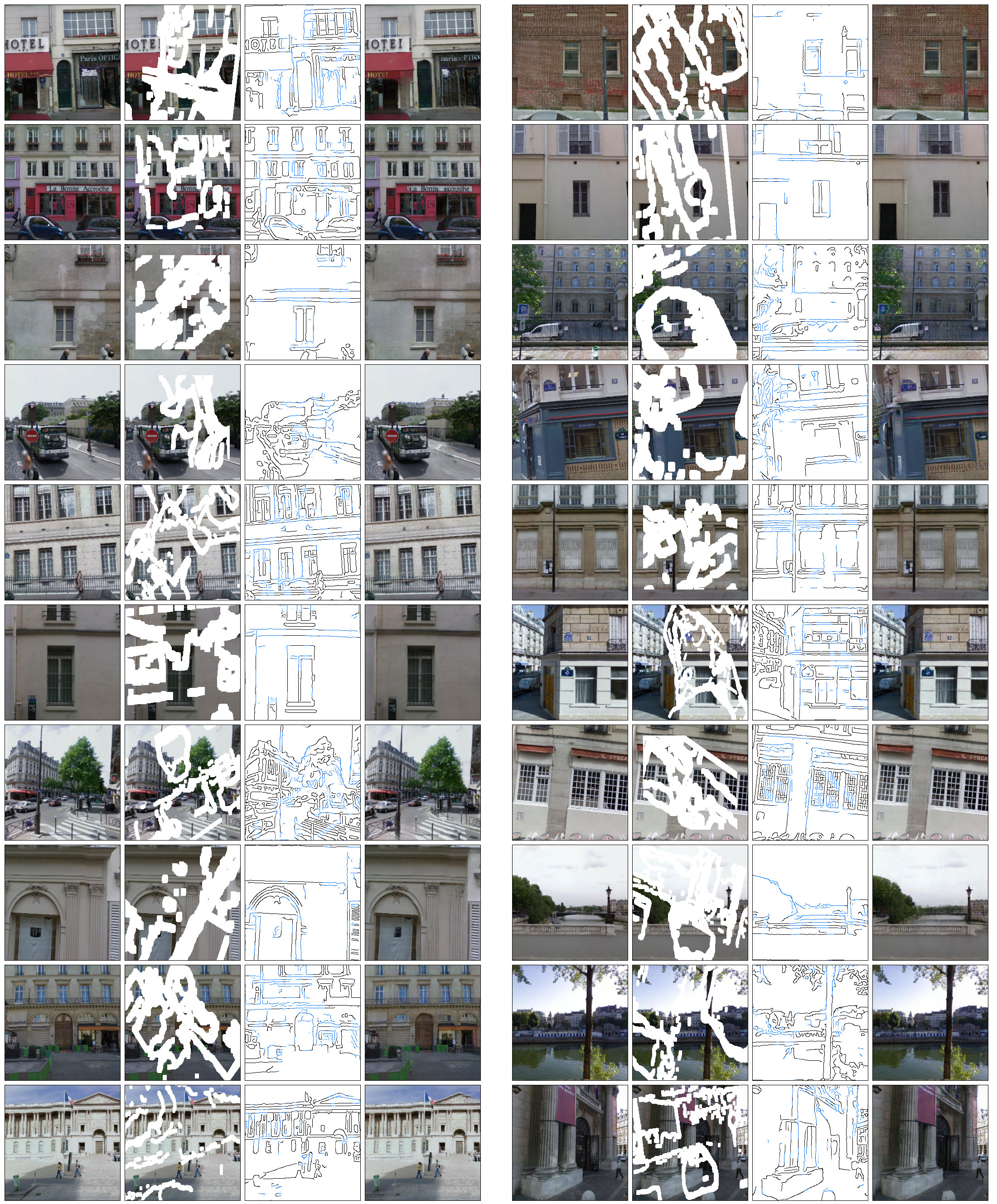}
			\caption*{(a)\hspace{50pt}(b)\hspace{50pt}(c)\hspace{50pt}(d)\hspace{65pt}(a)\hspace{50pt}(b)\hspace{50pt}(c)\hspace{50pt}(d)}
		\end{subfigure}
		\caption{Sample of results with Paris StreetView dataset. Images are best viewed in color. (a) Original Image. (b) Input Image. (c) Generated Edges (Blue) and Ground Truth Edges (Black). (d) Generated Result.}
		\label{fig:psv}
	\end{figure*}

	Tables \ref{tab:celeb} and \ref{tab:psv} shows the performance of our model compared to existing methods over the datasets CelebA and Paris StreetView respectively. Our method produces noticeably better results. Note that we did not include values for PConv over the CelebA and Paris StreetView datasets as the source code was not available at the time of writing. Figures \ref{fig:bar_celeb}, and \ref{fig:bar_psv} display these results graphically. Additional inpainting results of our proposed model are shown in figures \ref{fig:celeb}, \ref{fig:places}, and \ref{fig:psv}. \\\\
	Our source code, pre-trained models, and more results are available at:\\
	\small\url{https://github.com/knazeri/edge-connect}

\newpage
\section{Alternative Edge Generating Systems} \label{app:hed}

\begin{figure}[htb!]
		\centering
		\begin{subfigure}[c]{1\linewidth}
		\centering
			\includegraphics[width=1\linewidth]{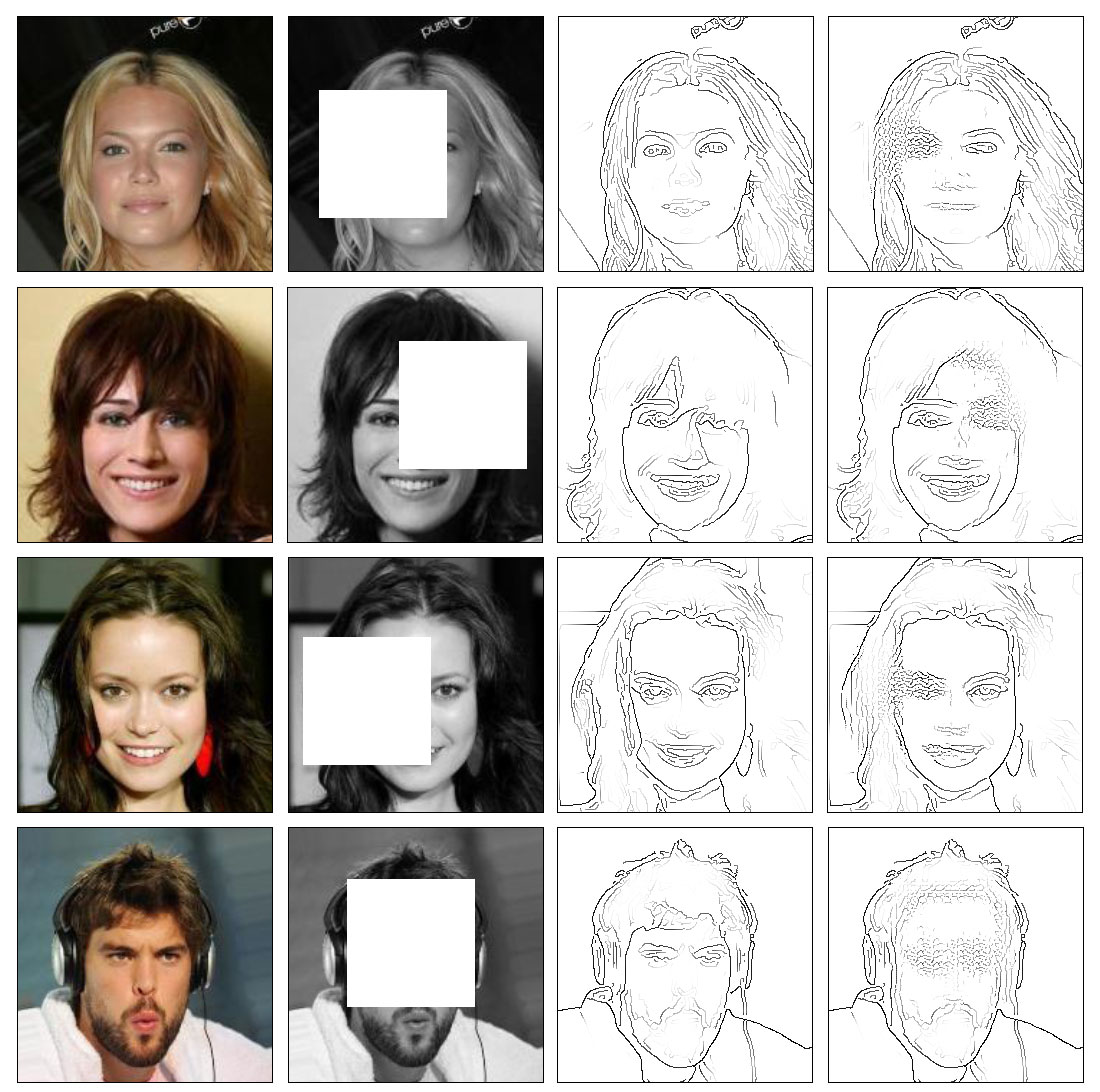}
			\caption*{(a)\hspace{47pt}(b)\hspace{47pt}(c)\hspace{47pt}(d)}
		\end{subfigure}
		\caption{Generated edges by $G_1$ trained using hybrid (HED$\odot$Canny) edges. Images are best viewed in color. (a) Original Image. (b) Image with Masked Region. (c) Ground Truth Edges. (d) Generated Edges.}
		\label{fig:fail_hed}
	\end{figure}

	In drawing, an edge is a boundary that separates two areas.	A thick line brings the shape forward thin line indicates a plane receding into the background. In other words edges create a sense of distance and are not just about lines. Here we use HED \cite{xie2015holistically} as an alternative edge detection system. Edges produced with HED, are of varying thickness, and pixels can have intensities ranging between 0 and 1. We noticed that it is possible to create edge maps that look eerily similar to human sketches by performing element-wise multiplication on Canny and HED edge maps. We compare the quantitative results between Canny and a combination of HED and Canny edges (\emph{i.e.} HED$\odot$Canny). Generated images based on the combined edges gave the best performance. However, our generator $G_1$ is unable to generate these type of edges accurately during training. Table \ref{tab:hed} shows $G_1$ trained on HED$\odot$Canny had the poorest performance out of all methods despite its ground truth counterpart achieving the best performance. These results suggest that better edge detectors result in better inpainting, however, effectively drawing those edges remains an open question in our research. Figure \ref{fig:fail_hed} shows the results of $G_1$ trained using hybrid edges.
	\newpage
	\begin{table}[htb!]
		\centering
		\def\arraystretch{1.2}
		\begin{tabular}{c|c*{4}{|>{\centering\arraybackslash}p{.1\linewidth}}}
			\multicolumn{2}{r|}{~} & \multicolumn{2}{c|}{Hybrid} & \multicolumn{2}{c}{Canny} \\ \cline{3-6}
			\multicolumn{2}{r|}{\textbf{Mask}} & \small{$G_1$} & \small{GT} & \small{$G_1$} & \small{GT} \\ \hhline{=*{4}{=|}=}
			\multirow{6}{*}{\rotatebox{90}{$\ell_1$ (\%)$^{\dagger}$}}
			& \small{0-10\%} & 0.31 & 0.23 & 0.29 & 0.25 \\ \cline{2-6}
			& \small{10-20\%} & 0.79 & 0.55 & 0.76 & 0.59 \\ \cline{2-6}
			& \small{20-30\%} & 1.42 & 0.93 & 1.38 & 1.00 \\ \cline{2-6}
			& \small{30-40\%} & 2.19 & 1.35 & 2.13 & 1.45 \\ \cline{2-6}
			& \small{40-50\%} & 3.10 & 1.82 & 3.03 & 1.97 \\ \cline{2-6}
			& \small{50-60\%} & 4.95 & 2.61 & 4.89 & 2.88 \\ \hhline{*{5}{=|}=}
			\multirow{6}{*}{\rotatebox{90}{SSIM$^{\star}$}}
			& \small{0-10\%} & 0.985 & 0.990 & 0.985 & 0.988 \\ \cline{2-6}
			& \small{10-20\%} & 0.959 & 0.978 & 0.961 & 0.972 \\ \cline{2-6}
			& \small{20-30\%} & 0.926 & 0.959 & 0.928 & 0.951 \\ \cline{2-6}
			& \small{30-40\%} & 0.886 & 0.940 & 0.890 & 0.930 \\ \cline{2-6}
			& \small{40-50\%} & 0.841 & 0.920 & 0.846 & 0.906 \\ \cline{2-6}
			& \small{50-60\%} & 0.767 & 0.891 & 0.771 & 0.872 \\ \hhline{*{5}{=|}=}
			\multirow{6}{*}{\rotatebox{90}{PSNR$^{\star}$}}
			& \small{0-10\%} & 39.24 & 42.43 & 39.60 & 41.77 \\ \cline{2-6}
			& \small{10-20\%} & 33.26 & 37.48 & 33.51 & 36.81 \\ \cline{2-6}
			& \small{20-30\%} & 29.80 & 34.65 & 30.02 & 34.00 \\ \cline{2-6}
			& \small{30-40\%} & 27.21 & 32.59 & 27.39 & 31.92 \\ \cline{2-6}
			& \small{40-50\%} & 25.12 & 30.87 & 25.28 & 30.21 \\ \cline{2-6}
			& \small{50-60\%} & 22.03 & 28.49 & 22.11 & 27.68 \\ \hhline{*{5}{=|}=}
			\multirow{6}{*}{\rotatebox{90}{FID$^{\dagger}$}}
			& \small{0-10\%} & 0.22 & 0.11 & 0.20 & 0.13 \\ \cline{2-6}
			& \small{10-20\%} & 0.56 & 0.24 & 0.53 & 0.31 \\ \cline{2-6}
			& \small{20-30\%} & 1.13 & 0.41 & 1.08 & 0.57 \\ \cline{2-6}
			& \small{30-40\%} & 1.90 & 0.61 & 1.80 & 0.88 \\ \cline{2-6}
			& \small{40-50\%} & 2.99 & 0.83 & 2.82 & 1.25 \\ \cline{2-6}
			& \small{50-60\%} & 5.67 & 1.14 & 5.30 & 1.79 \\ \hline
		\end{tabular}
		\caption{Comparison of quantitative results between Hybrid (HED$\odot$Canny) and Canny edges over CelebA. Statistics are shown for generated edges ($G_1$) and ground truth edges (GT). $^\dagger$Lower is better. $^\star$Higher is better.}
		\label{tab:hed}
	\end{table}

	\begin{table*}[t!]
		\parbox{.48\linewidth}
		{
			\def\arraystretch{1.3}
			\centering
			\begin{tabular}{c|c*{4}{|>{\centering\arraybackslash}p{.12\linewidth}}}
				\multicolumn{2}{r|}{\textbf{Mask}} & \small{CA} & \small{GLCIC} & \small{Ours} & \small{Canny} \\ \hhline{=*{4}{=|}=}
				\multirow{7}{*}{\rotatebox{90}{$\ell_1$ (\%)$^{\dagger}$}}
				& \small{0-10\%} & 1.33 & 0.91 & \textbf{0.29} & 0.25 \\ \cline{2-6}
				& \small{10-20\%} & 2.48 & 2.53 & \textbf{0.76} & 0.59 \\ \cline{2-6}
				& \small{20-30\%} & 3.98 & 4.67 & \textbf{1.38} & 1.00 \\ \cline{2-6}
				& \small{30-40\%} & 5.64 & 6.95 & \textbf{2.13} & 1.45 \\ \cline{2-6}
				& \small{40-50\%} & 7.35 & 9.18 & \textbf{3.03} & 1.97 \\ \cline{2-6}
				& \small{50-60\%} & 9.21 & 11.21 & \textbf{4.89} & 2.88 \\ \cline{2-6}
				& \small{Fixed} & 2.80 & 3.83 & \textbf{2.39} & 1.34 \\ \hhline{*{5}{=|}=}
				\multirow{7}{*}{\rotatebox{90}{SSIM$^{\star}$}}
				& \small{0-10\%} & 0.947 & 0.947 & \textbf{0.985} & 0.988 \\ \cline{2-6}
				& \small{10-20\%} & 0.888 & 0.865 & \textbf{0.961} & 0.972 \\ \cline{2-6}
				& \small{20-30\%} & 0.819 & 0.773 & \textbf{0.928} & 0.951 \\ \cline{2-6}
				& \small{30-40\%} & 0.750 & 0.689 & \textbf{0.890} & 0.930 \\ \cline{2-6}
				& \small{40-50\%} & 0.678 & 0.609 & \textbf{0.846} & 0.906 \\ \cline{2-6}
				& \small{50-60\%} & 0.614 & 0.560 & \textbf{0.771} & 0.872 \\ \cline{2-6}
				& \small{Fixed} & 0.882 & 0.847 & \textbf{0.891} & 0.944 \\ \hhline{*{5}{=|}=}
				\multirow{7}{*}{\rotatebox{90}{PSNR$^{\star}$}}
				& \small{0-10\%} & 31.16 & 30.24 & \textbf{39.60} & 41.77 \\ \cline{2-6}
				& \small{10-20\%} & 25.32 & 24.09 & \textbf{33.51} & 36.81 \\ \cline{2-6}
				& \small{20-30\%} & 22.09 & 20.71 & \textbf{30.02} & 34.00 \\ \cline{2-6}
				& \small{30-40\%} & 19.94 & 18.50 & \textbf{27.39} & 31.92 \\ \cline{2-6}
				& \small{40-50\%} & 18.41 & 17.09 & \textbf{25.28} & 30.21 \\ \cline{2-6}
				& \small{50-60\%} & 17.18 & 16.24 & \textbf{22.11} & 27.68 \\ \cline{2-6}
				& \small{Fixed} & 25.34 & 22.13 & \textbf{25.49} & 31.24 \\ \hhline{*{5}{=|}=}
				\multirow{7}{*}{\rotatebox{90}{FID$^{\dagger}$}}
				& \small{0-10\%} & 3.24 & 16.84 & \textbf{0.20} & 0.13 \\ \cline{2-6}
				& \small{10-20\%} & 13.12 & 58.74 & \textbf{0.53} & 0.31 \\ \cline{2-6}
				& \small{20-30\%} & 29.47 & 102.97 & \textbf{1.08} & 0.57 \\ \cline{2-6}
				& \small{30-40\%} & 47.55 & 136.47 & \textbf{1.80} & 0.88 \\ \cline{2-6}
				& \small{40-50\%} & 68.40 & 163.95 & \textbf{2.82} & 1.25 \\ \cline{2-6}
				& \small{50-60\%} & 76.70 & 167.07 & \textbf{5.30} & 1.79 \\ \cline{2-6}
				& \small{Fixed} & 1.90 & 25.21 & \textbf{1.90} & 0.74 \\ \hline
			\end{tabular}
			\caption{Comparison of quantitative results over CelebA with CA \cite{yu2018generative}, GLCIC \cite{iizuka2017globally}, Ours (end-to-end), Ours with Canny edges ($G_2$ only). The best result of each row is boldfaced except for Canny. $^\dagger$Lower is better. $^\star$Higher is better.}
			\label{tab:celeb}
		}
		\hspace{1cm}
		\parbox{.48\linewidth}
		{
			\vspace{.05\linewidth}
			\def\arraystretch{1.3}
			\begin{tabular}{c|c*{4}{|>{\centering\arraybackslash}p{.12\linewidth}}}
				\multicolumn{2}{r|}{\textbf{Mask}} & \small{CA} & \small{GLCIC} & \small{Ours} & \small{Canny} \\ \hhline{=*{4}{=|}=}
				\multirow{7}{*}{\rotatebox{90}{$\ell_1$ (\%)$^{\dagger}$}}
				& \small{0-10\%} & 0.75 & 0.86 & \textbf{0.43} & 0.40 \\ \cline{2-6}
				& \small{10-20\%} & 2.10 & 2.20 & \textbf{1.09} & 0.90 \\ \cline{2-6}
				& \small{20-30\%} & 3.80 & 3.86 & \textbf{1.91} & 1.48 \\ \cline{2-6}
				& \small{30-40\%} & 5.53 & 5.58 & \textbf{2.82} & 2.07 \\ \cline{2-6}
				& \small{40-50\%} & 7.23 & 7.34 & \textbf{3.94} & 2.77 \\ \cline{2-6}
				& \small{50-60\%} & 9.06 & 9.02 & \textbf{5.87} & 3.79 \\ \cline{2-6}
				& \small{Fixed} & 3.22 & 3.23 & \textbf{2.77} & 1.71 \\ \hhline{*{5}{=|}=}
				\multirow{7}{*}{\rotatebox{90}{SSIM$^{\star}$}}
				& \small{0-10\%} & 0.964 & 0.949 & \textbf{0.975} & 0.977 \\ \cline{2-6}
				& \small{10-20\%} & 0.905 & 0.878 & \textbf{0.938} & 0.949 \\ \cline{2-6}
				& \small{20-30\%} & 0.835 & 0.800 & \textbf{0.892} & 0.918 \\ \cline{2-6}
				& \small{30-40\%} & 0.766 & 0.724 & \textbf{0.842} & 0.886 \\ \cline{2-6}
				& \small{40-50\%} & 0.695 & 0.648 & \textbf{0.784} & 0.850 \\ \cline{2-6}
				& \small{50-60\%} & 0.625 & 0.588 & \textbf{0.700} & 0.804 \\ \cline{2-6}
				& \small{Fixed} & 0.847 & 0.840 & \textbf{0.860} & 0.909 \\ \hhline{*{5}{=|}=}
				\multirow{7}{*}{\rotatebox{90}{PSNR$^{\star}$}}
				& \small{0-10\%} & 32.45 & 30.46 & \textbf{36.31} & 37.38 \\ \cline{2-6}
				& \small{10-20\%} & 26.09 & 25.72 & \textbf{31.23} & 33.38 \\ \cline{2-6}
				& \small{20-30\%} & 22.80 & 22.90 & \textbf{28.26} & 31.04 \\ \cline{2-6}
				& \small{30-40\%} & 20.74 & 21.02 & \textbf{26.05} & 29.36 \\ \cline{2-6}
				& \small{40-50\%} & 19.35 & 19.66 & \textbf{24.20} & 27.85 \\ \cline{2-6}
				& \small{50-60\%} & 18.17 & 18.71 & \textbf{21.73} & 25.92 \\ \cline{2-6}
				& \small{Fixed} & 23.68 & 24.07 & \textbf{25.23} & 29.62 \\ \hhline{*{5}{=|}=}
				\multirow{7}{*}{\rotatebox{90}{FID$^{\dagger}$}}
				& \small{0-10\%} & 2.26 & 6.50 & \textbf{0.44} & 0.31 \\ \cline{2-6}
				& \small{10-20\%} & 9.10 & 18.77 & \textbf{1.20} & 0.68 \\ \cline{2-6}
				& \small{20-30\%} & 20.62 & 35.66 & \textbf{2.49} & 1.24 \\ \cline{2-6}
				& \small{30-40\%} & 34.31 & 53.53 & \textbf{4.35} & 1.89 \\ \cline{2-6}
				& \small{40-50\%} & 49.80 & 70.36 & \textbf{7.20} & 2.78 \\ \cline{2-6}
				& \small{50-60\%} & 55.78 & 69.95 & \textbf{13.98} & 4.12 \\ \cline{2-6}
				& \small{Fixed} & 7.26 & 7.18 & \textbf{4.57} & 3.24 \\ \hline
			\end{tabular}
			\caption{Comparison of quantitative results over Paris StreetView with CA \cite{yu2018generative}, GLCIC \cite{iizuka2017globally}, Ours (end-to-end), Ours with Canny edges ($G_2$ only). The best result of each row is boldfaced except for Canny. $^\dagger$Lower is better. $^\star$Higher is better.}
			\label{tab:psv}
		}
	\end{table*}

	

\end{document}